%% file: main.tex
\documentclass[conference]{IEEEtran}
\IEEEoverridecommandlockouts
\usepackage{cite}
\usepackage{amsmath,amssymb,amsfonts}
\usepackage{algorithmic}
\usepackage{graphicx}
\usepackage{textcomp}
\usepackage{xcolor}
\def\BibTeX{{\rm B\kern-.05em{\sc i\kern-.025em b}\kern-.08em
    T\kern-.1667em\lower.7ex\hbox{E}\kern-.125emX}}
    
\input{defs.tex}

\begin{document}

\title{Two-timescale Mechanism-and-Data-Driven Control for Aggressive Driving of Autonomous Cars\thanks{This work is supported by the National Key Research and Development Program of China under Grant 2018AAA0101601.}
}

\author{\IEEEauthorblockN{Yiwen Lu}
\IEEEauthorblockA{\textit{Dept. of Automation and BNRist} \\
\textit{Tsinghua University}\\
Beijing, China \\
luyw20@mails.tsinghua.edu.cn}
\and
\IEEEauthorblockN{Bo Yang}
\IEEEauthorblockA{\textit{Dept. of Automation and BNRist} \\
\textit{Tsinghua University}\\
Beijing, China \\
yang-b21@mails.tsinghua.edu.cn}
\and
\IEEEauthorblockN{Yilin Mo}
\IEEEauthorblockA{\textit{Dept. of Automation and BNRist} \\
\textit{Tsinghua University}\\
Beijing, China \\
ylmo@tsinghua.edu.cn}
}

\maketitle

\begin{abstract}
The control for aggressive driving of autonomous cars is challenging due to the presence of significant tyre slip. Data-driven and mechanism-based methods for the modeling and control of autonomous cars under aggressive driving conditions are limited in data efficiency and adaptability respectively. This paper is an attempt toward the fusion of the two classes of methods.
By means of a modular design that is consisted of mechanism-based and data-driven components, and aware of the two-timescale phenomenon in the car model, our approach effectively improves over previous methods in terms of data efficiency, ability of transfer and final performance. The hybrid mechanism-and-data-driven approach is verified on TORCS (The Open Racing Car Simulator). Experiment results demonstrate the benefit of our approach over purely mechanism-based and purely data-driven methods.
\end{abstract}


\input{intro.tex}
\input{problem.tex}
\input{method.tex}
\input{simulation.tex}

\bibliography{main.bib}

\end{document}

%% file: defs.tex
\usepackage{tikz}
\usepackage{tikzscale}
\usepackage{tkz-euclide}
\usepackage{subcaption}
\usepackage{caption}
\usepackage{pgfplots}
\usepackage{url}
\usepackage{tabularx,booktabs}

\usetikzlibrary{shapes}
\usetikzlibrary{matrix,arrows,fit,backgrounds,mindmap,plotmarks,decorations.pathreplacing}
\usetikzlibrary{shapes.multipart}

\definecolor{thupurple}{RGB}{0,0,0}

%% file: intro.tex
\section{Introduction}
\label{sec:intro}

In recent years, autonomous cars have demonstrated vast potential in terms of increasing road efficiency, reducing accidents and alleviating environmental impact~\cite{ryan2019future}.
The development of high~(L4) and full~(L5) driving automation has been put on the agenda worldwide, and improving the applicability of autonomous driving in complicated road conditions has drawn significant research attention~\cite{campbell2010autonomous,hubmann2017decision,talpaert2019exploring}.

This paper considers the control of autonomous cars under aggressive driving conditions. Subject to the difficulty of control under high speed and with the presence of significant tyre slip, existing practical driving automation schemes usually adopt conservative policies, and the driving efficiency has room for improvement. Therefore, aggressive autonomous driving scenes like high-speed off-road driving~\cite{Jeon2011a} and autonomous overtaking~\cite{Dixit2018} have been studied in recent years. Research into how to make the car execute aggressive driving behavior while ensuring safety would help us explore the boundary of autonomous driving, and provide technical insights toward high and full driving automation.

We adopt a modular approach to the modeling and control of autonomous cars. In particular, we split the car model into modules with fast and slow timescales, and propose a controller with mechanism-based and data-driven modules. The simplified mechanism for driving an autonomous car is i) the controller changes the steering angle and exerts torques to the wheels; ii) the tyres respond to the steering angle and the torques, yielding friction forces through tyre-road interaction; iii) the friction forces cause the pose of the car body to change. We observe that friction forces respond to the control inputs much faster than the pose of the car, because the inertia of a wheel is much smaller than that of the car body. In other words, the wheels correspond to the fast timescale, while the car body corresponds to the slow timescale, and this observation enables us to design an efficient two-timescale control scheme. Furthermore, we notice that the car body can be idealized as a rigid body, which can be readily modeled and controlled from first principles, while the tyre-road interaction characteristic is a highly nonlinear function between the slip ratio and the friction force that can only be described by an empirical formula called Pacejka Magic Formula~\cite{pacejka1997magic}, which involves parameters that are difficult to measure, necessitating data-driven modeling. Therefore, we include a mechanism-based module for driving the rigid body and a data-driven module for dealing with the tyre-road interaction in our control policy. This strategy demonstrates higher data efficiency than the purely data-driven scheme and better adaptability than the purely mechanism-based scheme.

Existing research papers have studied integrating mechanism-based and data-driven control methods in autonomous driving tasks from different aspects. Rosolia et al.~\cite{Rosolia2019} addresses the racing task by applying Model Predictive Control~(MPC) to track a reference trajectory that is iteratively learned on a lap-to-lap basis. However, it assumes a relatively accurate low-level dynamics model of the car is obtained \textit{a priori} through system identification, and suffers from limited generalization across different tyre-road interaction characteristics, which restricts its applicability in general high-performance driving tasks. Hewing et al.~\cite{Hewing_Liniger_Zeilinger_2018} models the car dynamics
as the sum of the nominal dynamics and a Gaussian Process (GP), which leverages the power of both mechanism and data, but it makes no distinction between mechanism-based and data-driven modules. By contrast, we adopt a modular design that marks a clear boundary between mechanism-based and data-driven components, and explicitly considers the two-timescale issue mentioned above. This modular design improves the explainability of control policy, and is suitable for the deployment over various aggressive driving scenarios.

To summarize, in this paper we propose a two-timescale control scheme for aggressive autonomous driving that combines mechanism-based and data-driven methods. We verify our method on both a simplified car model with tyre slip and TORCS racing simulator~\cite{wymann2000torcs}. Simulation results show that our proposed method can achieve significantly higher data efficiency than the purely data-driven method, and is more adaptive to various driving conditions than the purely mechanism-based method.

%% file: problem.tex
\section{Problem Formulation}\label{sec:problem}

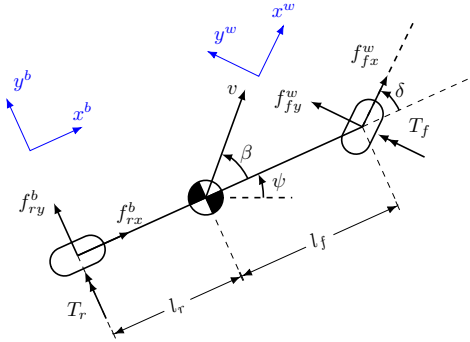
\begin{figure}[!htbp]
    \centering
    \resizebox{0.8\columnwidth}{!}{
        \hspace{-1cm}
        \input{figures/bike_model.tikz}
    }
    \caption{Illustration of car model}
    \label{fig:bike_model}
\end{figure}

We use the bicycle model with tyre slip and load transfer~\cite{Jeon2011a} for simulation and control, illustrated in Fig.~\ref{fig:bike_model}.

The state and input vectors are
\begin{align}
    \mathbf{X}=\left[x, y, \psi, \dot{x}, \dot{y}, \dot{\psi}, \omega_{f}, \omega_{r}\right]^\top, \mathbf{U}=\left[\delta, T_{f}, T_{r}\right]^\top,
    \label{eq:xu}
\end{align}
where $x,y$ are the coordinate of the car in a two-dimensional plane, $\psi$ is the heading angle of the car, $\dot{x}, \dot{y}, \dot{\psi}$ are the time derivatives of $x,y,\psi$; $\omega_f, \omega_r$ are the rotational speed of the front and rear wheels respectively, $\delta$ is the steering angle, and $T_f, T_r$ are the torques exerted on the front and rear wheels respectively. The dynamic equations are
\small
\begin{align}
    m\ddot x &= f_{fx}^w \cos(\psi + \delta) - f_{fy}^w \sin(\psi + \delta) + f_{rx}^b \cos\psi - f_{ry}^b \sin\psi, \label{eq:xddot}
    \\
    m\ddot y &= f_{fx}^w \sin(\psi + \delta) + f_{fy}^w \cos(\psi + \delta) + f_{rx}^b \sin\psi + f_{ry}^b \cos\psi,  \label{eq:yddot}
    \\
    I_z \ddot\psi &= \left( f_{fy}^w\cos\delta + f_{fx}^w \sin\delta \right) l_f - f_{ry}^b l_r, \label{eq:psiddot}
    \\
    I_f \dot{\omega}_f  &= T_f - f_{fx}^w r_f, 
    I_r \dot{\omega}_r  = T_r - f_{rx}^b r_r, \label{eq:omegadot}
\end{align}
\normalsize
where $m$ is the mass of the car, $I_z$ is the moment of inertia of the car body w.r.t. the $z$ axis, $I_f, I_r$ are the moments of inertia of the wheels, $r_f, r_r$ are the radii of the wheels, $l_f, l_r$ are the distances of the wheels to the center of mass of the car, and $f_{fx}^w, f_{fy}^w, f_{rx}^b, f_{ry}^b$ are the friction forces. The subscripts `f', `r' refer to ``front'' and ``rear'' respectively, and the superscripts `w' and `b' refer to ``front wheel frame'' and ``body frame'' respectively.
It shall be noticed that the use of these two coordinate systems facilitates the decomposition of body and wheel dynamics.
The forces are determined by
\begin{align*}
    f_{fx}^w = \mu_{fx}f_{fz},   f_{fy}^w = \mu_{fy}f_{fz},  
f_{rx}^b = \mu_{rx}f_{rz},   f_{ry}^b = \mu_{ry}f_{rz},     
\end{align*}
where $\mu_{fx}, \mu_{fy}, \mu_{rx}, \mu_{ry}$ are the friction coefficients, and $f_{fz}, f_{rz}$ are the normal forces. The friction coefficients are described by Pacejka Magic Formula~\cite{pacejka1997magic}:
\begin{align}
    \mu_{ij} = - \frac{s_{ij}}{s_{i}} D \sin\left( C \arctan(Bs_i) \right) \; \left(i \in \{f,r\},\; j \in \{x,y\}\right),
    \label{eq:mf}
\end{align}
where $B,C,D$ are parameters that vary with the tyre-road interaction properties, and $s_i, s_{ij}$ are slip ratios that can be computed from relative speeds between the wheels and the road:
\begin{align}
    & s_i = \sqrt{s_{ix}^2 + s_{iy}^2}, \quad \left(i \in \{f,r\}\right), \label{eq:def_s} \\
    & s_{fx} = \frac{v_{fx}^w - \omega_f r_f}{\omega_f r_f}, s_{fy} = \frac{v_{fy}^w}{\omega_f r_f}, \label{eq:def_sf}\\
    &
    s_{rx} = \frac{v_{rx}^b - \omega_r r_r}{\omega_r r_r}, s_{ry} = \frac{v_{ry}^b}{\omega_r r_r}, \label{eq:def_sr}
    \\
        & v=\sqrt{\dot{x}^{2}+\dot{y}^{2}},  \beta=\arctan \frac{\dot{y}}{\dot{x}}-\psi, \\
    & v_{f x}^w=v \cos (\beta-\delta)+\dot{\psi} l_{f} \sin \delta,  v_{rx}^b = v\cos\beta, \\
    & v_{f y}^w=v \sin (\beta-\delta)+\dot{\psi} l_{f} \cos \delta,  v_{ry}^b = v\sin\beta - \dot\psi l_r.  \label{eq:vfyw}
\end{align}
Finally, the normal forces $f_{fz}, f_{rz}$ can be determined as:
\begin{align*}
        f_{fz} = \frac{l_r - \mu_{rx}h}{l_f + l_r + \left(\mu_{f x} \cos \delta-\mu_{f y} \sin \delta-\mu_{r x}\right)h}mg,\\
        f_{rz} = \frac{l_f + \left( \mu_{f x} \cos \delta-\mu_{f y} \sin \delta \right)h}{l_f + l_r + \left(\mu_{f x} \cos \delta-\mu_{f y} \sin \delta-\mu_{r x}\right)h}mg.
\end{align*}

The goal of the controller is to drive the car to optimize some certain performance metrics subject to safety constraints, without prior knowledge of the tyre-road interaction parameters $B,C,D$. We assume all the state variables can be directly measured. Mathematically, the controller should approximately solve the optimal control problem
\begin{align}
\begin{aligned}
    & \min_{\mathbf{X}_{0:T-1},\mathbf{U}_{0:T-1}} J\left(\mathbf{X}_{0:T-1}, \mathbf{U}_{0:T-1}\right), \\
    & \text{s.t. } \mathbf{X}_{t+1} = f\left(\mathbf{X}_t, \mathbf{U}_t\right), \mathbf{X}_t \in \mathcal{X}, \mathbf{U}_t \in \mathcal{U}, \mathbf{X}_0 = \mathbf{X}_{\text{init}},
\end{aligned}
\label{eq:problem}
\end{align}
where $\mathbf{X}_t, \mathbf{U}_t$ are state and input vectors respectively, as defined in~\eqref{eq:xu}, $T$ is the control horizon, $J$ is the performance metrics, $f$ is the dynamics function, $\mathcal{X}, \mathcal{U}$ are the constraint sets on the state and input respectively, and $\mathbf{X}_{\text{init}}$ is the initial state. In particular, we consider the following tasks in this paper: \begin{itemize}
    \item Low-level task - path tracking: track a manually specified path with high speed and large curvature in an infinite plane. In this task, $J$ is the tracking error, $f$ is the discretized system of the dynamics described in~\eqref{eq:xddot}-\eqref{eq:omegadot}, $\mathcal{U}$ is the maneuver limit of the car, and there is no additional constraint on the state.
    \item High-level task - racing: race on TORCS platform with tracks in different shapes. In this task, $J$ is the lap time, $f$ is a high-fidelity simulation model more complicated than the one described in this section, $\mathcal{U}$ is the maneuver limit of the car, and $\mathcal{X}$ represent the road boundaries.
\end{itemize}

%% file: figures/bike_model.tikz
\usetikzlibrary{calc,angles,quotes,arrows.meta}

\tikzset{pill/.style={minimum width=1cm,minimum height=5mm,rounded corners=2.5mm,draw},
reactor/.style={circle,draw,minimum size=6mm,path picture={
\draw (-3mm,0) -- (3mm,0) (0,-3mm) -- (0,3mm);
\fill (0,0) -- (3mm,0) arc(0:-90:3mm) -- cycle;
\fill (0,0) -- (-3mm,0) arc(180:90:3mm) -- cycle;
}}}

\begin{tikzpicture}
\coordinate (O) (0, 0);
\path (O) -- (9.5,4.3) coordinate[pos=0.28] (F1) coordinate[pos=0.8] (F2) coordinate (TR);
\draw[thick] (F1)  -- (F2) node[pos=0.45,sloped,reactor] (M){~}
node[pos=0,sloped,pill]{};
\draw[dashed] (F2) -- (TR);
\draw[thick,-latex] (F1) -- ($(F1)!1cm!0:(F2)$)
node[above]{$f_{rx}^b$};
\draw[thick,-latex] (F1) -- ($(F1)!1cm!90:(F2)$)
node[left]{$f_{ry}^b$};
\draw[thick,dashed] (F2) -- ++ (64:2) coordinate(H)  node[pos=0,sloped,pill,solid]{}
pic ["$\delta$",draw,solid,-latex,angle radius=0.7cm,angle eccentricity=1.3] {angle = TR--F2--H};
\draw[thick,-latex] (F2) -- ($(F2)!1cm!0:(H)$)
node[above left]{$f_{fx}^w$};
\draw[thick,-latex] (F2) -- ($(F2)!1cm!90:(H)$)
node[left]{$f_{fy}^w$};
\path (F2) -- ($(F2)!1.5cm!270:(H)$) coordinate[pos=0.2] (F21) coordinate[pos=0.8] (F22) ;
\draw[{Bar}{Latex}-{Latex}{Bar}] ($(F1)!1.5cm!270:(M)$) coordinate (l1) -- 
($(M)!1.5cm!270:(F2)$) coordinate (l2) node[midway,sloped,fill=white]{$l_r$};
\draw[{Bar}{Latex}-{Latex}{Bar}] (l2) -- 
($(F2)!1.5cm!90:(M)$) coordinate (l3) node[midway,sloped,fill=white]{$l_f$};
\draw[dashed] (F1) -- (l1) coordinate[pos=0.2] (F11) coordinate[pos=0.8] (F12) ;
\draw[dashed] (M.center) -- (l2);
\draw[dashed] (F2) -- (l3) ;
\draw[thick,->>,>=latex] (F12) -- (F11) node[midway,below left] {$T_r$};
\draw[thick,->>,>=latex] (F22) -- (F21) node[midway,above right] {$T_f$};

\draw[thick,dashed] (M.center) -- ++(0:1.5) coordinate (M1) node{}
pic ["$\psi$",draw,solid,-latex,angle radius=1cm,angle eccentricity=1.3] {angle = M1--M--F2};

\draw[thick,-latex] (M.center) -- ++(70:2) coordinate (V) node[left]{$v$}
pic ["$\beta$",draw,solid,-latex,angle radius=0.8cm,angle eccentricity=1.3] {angle = F2--M--V};

\draw[blue,{Latex}-{Latex}] ($(F1)!3cm!90:(F2)$) node [above right] {$y^b$} coordinate (Yb) -- ($(F1)!2cm!90:(F2)$) coordinate (Ob) -- ($(Ob)!1cm!-90:(Yb)$) node [above] {$x^b$};
\draw[blue,{Latex}-{Latex}] ($(F2)!3cm!90:(H)$) node [above right] {$y^w$} coordinate (Yw) -- ($(F2)!2cm!90:(H)$) coordinate (Ow) -- ($(Ow)!1cm!-90:(Yw)$) node [above] {$x^w$};

\end{tikzpicture}

%% file: method.tex
\section{Controller Design}

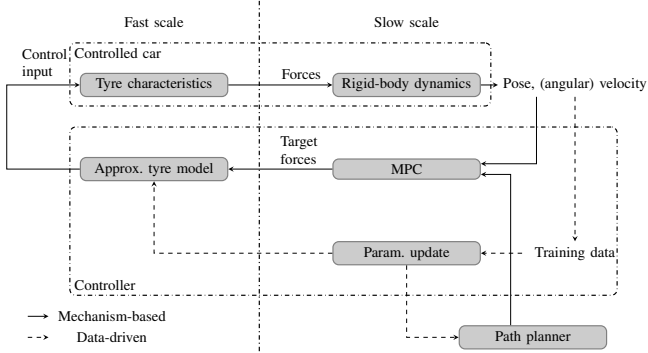
\begin{figure}[!htbp]
    \centering
    \resizebox{\columnwidth}{!}{
        \input{figures/framework.tikz}
    }
    \caption{Schematic diagram of closed-loop system}
    \label{fig:framework}
\end{figure}

In this section, we first illustrate the two-timescale phenomenon in car model using an example, which reveals the rationale behind our modular design. Afterwards, we describe the mechanism-based and data-driven components of our controller. A schematic diagram of the closed-loop system of the car driven by our controller is shown in Fig.~\ref{fig:framework}.

\subsection{Two-timescale phenomenon in car model}

In theory, it is possible to find an approximate solution to the optimal control problem~\eqref{eq:problem} using Model Predictive Control~(MPC), which can be solved numerically using a nonlinear solver such as Ipopt~\cite{Biegler_Zavala_2009}. However, we next illustrate with an example that this is impractical.

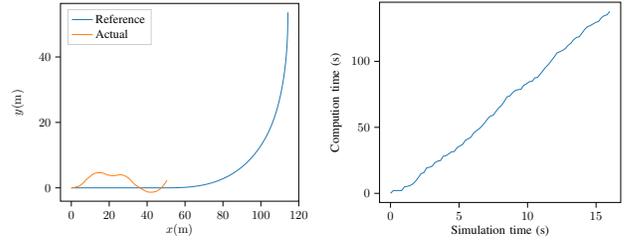
\begin{figure}[!htbp]
    \centering
    \subcaptionbox{Comparison of reference trajectory and actual trajectory\label{fig:ref_traj}}{\resizebox{0.465\columnwidth}{!}{\input{figures/ill_traj.tikz}}}
    \subcaptionbox{Computation time}{\resizebox{0.45\columnwidth}{!}{\input{figures/ill_time.tikz}}}
    \caption{Result of directly applying MPC}
    \label{fig:mpc_ill}
\end{figure}

In the following example, we instantiate the simulation model in Section~\ref{sec:problem} with the geometric and mechanical parameters of a Mercedes CLS 63 AMG car\footnote{Data available from \url{https://www.teoalida.com/cardatabase/}.} and the tyre-road interaction parameters of a typical tarmac road ($B=10,C=1.9,D=1$). We choose the frequency of the controller to be $10\mathrm{Hz}$ and discretize the dynamics model~\eqref{eq:xddot}-\eqref{eq:omegadot} with fourth-order Runge-Kutta method. We apply MPC to track the reference trajectory shown in Fig.~\ref{fig:ref_traj}. It can be observed from Fig.~\ref{fig:mpc_ill} that the controller cannot achieve reasonable tracking performance even though the computation is already slower than real time.

To analyze why the direct application of MPC fails, we inspect the Jacobian matrix $A(t)=\left.\frac{\partial \left[ \dot{\psi}, \ddot{x}, \ddot{y}, \ddot{\psi} , \dot{\omega}_f, \dot{\omega}_r \right]^\top}{\partial \left[ {\psi}, \dot{x}, \dot{y}, \dot{\psi} , {\omega}_f, {\omega}_r \right]^\top}\right|_{\mathbf{X}=\mathbf{X}(t), \mathbf{U}=\mathbf{U}(t)}$ at a particular time step $t$ along the reference trajectory. We have
\small
\begin{align*}
    A(t) \approx\left[\begin{array}{cccccc}
0 & 0 & 0 & 1 & 0 & 0 \\
0 & -200 & 0 & 0 & 32.0 & 33.6 \\
1.05 & 0 & -194 & 7.91 & 0 & 0 \\
0 & 0 & 11.3 & -567 & 0 & 0 \\
0 & 1.4\mathrm{e}4 & 0 & 0 & -4.6 \mathrm{e}3 & -4.29 \\
0 & 1.8\mathrm{e}4 & 0 & 0 & -208 & -5.7 \mathrm{e}3
\end{array}\right],
\end{align*}
\normalsize
of which the eigenvalue with largest absolute value is $\bar{\lambda} \approx-1.16 \times 10^{3}$, and the eigenvalue of the smallest absolute value is $\underline{\lambda} \approx 2.16 \times 10^{-5}$. We have $|\bar{\lambda}| /|\underline{\lambda}| \approx 5.37 \times 10^{7}$, which indicates $A(t)$ is a highly ill-conditioned matrix.
It is known from the theory of numerical solutions for ordinary differential equations~\cite{Lambert_1991} that the differential equation $\dot{\mathbf{X}} = \dot{\mathbf{X}}\left(\mathbf{X}, \mathbf{U}\right)$ has a high stiffness ratio, and therefore, explicit solvers like Euler and Runge-Kutta are numerically unstable for this differential equation unless the step size is taken extremely small.
Correspondingly, the discretization $\mathbf{X}_{t+1}=f\left(\mathbf{X}_{t}, \mathbf{U}_{t}\right)$ in the MPC formulation~\eqref{eq:problem} is very inaccurate under normal step sizes, posing challenge to the implementation of the controller.

From a physical point of view, the inertia of the wheels are significantly smaller than that of the car body, and therefore the wheels respond to changes much faster than the car body. In the matrix $A(t)$ demonstrated above, the elements with largest magnitudes are $\frac{\partial \dot{\omega}_{f}}{\partial \dot{x}}, \frac{\partial \dot{\omega}_{r}}{\partial \dot{x}}, \frac{\partial \dot{\omega}_{f}}{\partial \omega_{f}}, \frac{\partial \dot{\omega}_{r}}{\partial \omega_{r}}$, which means the change of wheel rotational speeds is very sensitive to the current rotational speed and the longitudinal speed of the car. By contrary, the pose of the car body is not very sensitive to the instantaneous states and inputs, and the inputs need to be accumulated over time in order to drive the car to the desired pose. The above observation can be summarized as a two-timescale phenomenon, which we explicitly consider in our controller design. In particular, at the slow timescale, we design a controller to make the car body track the reference trajectory assuming the friction forces $f_{fx}, f_{fy}, f_{rx}, f_{ry}$ can be directly controlled, while at the fast timescale, we solve for the actual control inputs that yield the desired forces assuming the motion of the car body is constant. This modular design is described in detail in the following subsections.

\vspace{-0.15cm}

\subsection{Mechanism-based controller}

At the slow timescale, we focus our attention on the car body, and use MPC to solve for the desired friction forces based on rigid-body mechanics principles. The state vector for this MPC is the pose of the car body along with its derivative. Next we determine the input vector for the MPC.
Since the forces should be independent of the steering angle of the front wheel $\delta$ at the slow timescale, we cast $f_{fx}^w, f_{fy}^w$ into the body frame through the identity
\begin{align*}
\left[\begin{array}{c}
f_{f x}^{b} \\
f_{f y}^{b}
\end{array}\right]=\left[\begin{array}{cc}
\cos \delta & -\sin \delta \\
\sin \delta & \cos \delta
\end{array}\right]\left[\begin{array}{l}
f_{f x}^{w} \\
f_{f y}^{w}
\end{array}\right]
\end{align*}
as illustrated in Fig.~\ref{fig:bike_model}.
To summarize, the state and input vectors for the MPC are
\begin{align*}
    \mathbf{q}=\left[x, y, \psi, \dot{x}, \dot{y}, \dot{\psi}\right]^\top, \mathbf{f}=\left[ f_{fx}^b, f_{rx}^b, f_{fy}^b, f_{ry}^b \right]^ \top.
\end{align*}

The controller solves the tracking problem
\begin{align}
\begin{aligned}
    & \min_{\mathbf{q}_{0:T-1},\mathbf{f}_{0:T-1}} \sum_{t=0}^{T-1} \left(\mathbf{q}_t - \mathbf{q}^{\text{ref}}_t \right)^\top \mathbf{Q} \left(\mathbf{q}_t - \mathbf{q}^{\text{ref}}_t \right) + \mathbf{f}_t^\top \mathbf{R} \mathbf{f}_t , \\
    & \text{s.t. } \mathbf{q}_{t+1} = f\left(\mathbf{q}_t, \mathbf{f}_t\right), \underline{\mathbf{f}} \leq \mathbf{f}_t \leq \bar{\mathbf{f}}, \mathbf{q}_0 = \mathbf{q}_{\text{current}},
\end{aligned}
\label{eq:mpc}
\end{align}
at each step and takes $\mathbf{f}_0$ for further processing as is clarified later in this subsection. In~\eqref{eq:mpc}, $T$ is the horizon for the MPC, $\mathbf{Q},\mathbf{R}$ are fixed positive-definite weight matrices, $f$ is the discretized system of the rigid-body equations~\eqref{eq:xddot}-\eqref{eq:psiddot}, and $\underline{\mathbf{f}},\bar{\mathbf{f}}$ are approximate lower and upper bounds for the corresponding forces that are specified by design choices. We assume that reference trajectory $\left\{ \mathbf{q}^{\text{ref}}_t \right\}$ is available to MPC, either manually specified or provided by a task-specific planner module.

At the fast timescale, we focus our attention on the wheels, in order to translate the desired friction forces $\mathbf{f}_0$ obtained from the MPC at the slow timescale into actual control inputs $\mathbf{U}$. In particular, we assume $\mathbf{q}$ which describes the pose and motion of the car body is constant, in which situation we choose the steering angle and torques to make the actual friction forces as close to  $\mathbf{f}_0$ as possible. We will proceed in two steps. Firstly, we convert the desired friction forces into the steering angle $\delta$ and the slip ratios $s_{fx}, s_{fy}, s_{rx}, s_{ry}$. Secondly, we determine the torques that would lead to the desired slip ratios.

We exploit the coupling between lateral and longitudinal slip to determine the desired slip ratios. From~\eqref{eq:mf} and~\eqref{eq:def_sr} we have
\begin{align*}
    \frac{s_{r x}}{s_{r y}}=\frac{\mu_{r x}}{\mu_{r y}}=\frac{f_{r x}^{b}}{f_{r y}^{b}}, \frac{s_{r x}+1}{s_{r y}}=\frac{v_{r x}^{b}}{v_{r y}^{b}},
\end{align*}
and by taking the difference of the above to equalities, we have
\begin{align*}
    s_{r y}=\left(\frac{v_{r x}^{b}}{v_{r y}^{b}}-\frac{f_{r x}^{b}}{f_{r y}^{b}}\right)^{-1}, s_{r x}=\frac{f_{r x}^{b}}{f_{r y}^{b}}\left(\frac{v_{r x}^{b}}{v_{r y}^{b}}-\frac{f_{r x}^{b}}{f_{r y}^{b}}\right)^{-1}.
\end{align*}
According to the ``constant motion'' assumption for the fast timescale, we can substitute the measured velocity components $v_{rx}^b, v_{ry}^b$ into the above equation to obtain the desired $s_{rx},s_{ry}$. Similarly, for the front wheel we have
\begin{align*}
    &\frac{s_{f x}}{s_{f y}}=\frac{\mu_{f x}}{\mu_{f y}}=\frac{f_{f x}^{w}}{f_{f y}^{w}}=\frac{\cos \delta f_{f x}^{b}+\sin \delta f_{f y}^{b}}{-\sin \delta f_{f x}^{b}+\cos \delta f_{f y}^{b}}, \\
    &\frac{s_{f x}+1}{s_{f y}}=\frac{v_{f x}^{w}}{v_{f y}^{w}}=\frac{v \cos (\beta-\delta)+\dot{\psi} l_{f} \sin \delta}{v \cos \beta},
\end{align*}
from which we can represent $s_{fx}, s_{fy}$ as functions of $\delta$. To determine the steering angle $\delta$, we impose the constraint that the total slip on the front wheel $s_{f}(\delta)=\sqrt{s_{f x}(\delta)^{2}+s_{f y}(\delta)^{2}}$ should yield the total desired friction force on the front wheel, i.e., $\delta$ should satisfy
the equation
\begin{align}
    D \sin \left(C \arctan \left(B s_f(\delta) \right)\right) = \frac{\sqrt{\left(f_{f x}^{b}\right)^{2}+\left(f_{f y}^{b}\right)^{2}}}{f_{f z}}.
    \label{eq:delta}
\end{align}
By substituting the unknown tyre-road interaction parameters $B,C,D$ by their estimated values provided by the data-driven module described in the next subsection, and $f_{fz}$ by its measured value, we can solve the nonlinear equation~\eqref{eq:delta} numerically to obtain $\delta$, and hence $s_{fx}, s_{fy}$.

To determine the torques, we first compute how much change in wheel rotational speeds $\omega_f, \omega_r$ we need in order to achieve the desired slip ratios, and then apply linear approximation to the wheel dynamics~\eqref{eq:omegadot} to obtain $T_f, T_r$. To clarify, let us take the front wheel as an example, and linear approximation of~\eqref{eq:omegadot} results in
\begin{align}
    {\omega}_{f}(t)\approx \omega_{f}(0)+\frac{t}{I_{f}}\left(T_{f}-f_{f x}^{w} r_{f}\right).\label{eq:lin_approx}
\end{align}
Let the control interval be $\Delta t$, then under the above linear approximation, the average $\omega_f$ over an interval is $\omega_f(\Delta t / 2)$. We command the longitudinal slip produced by the average $\omega_f$ to equal the desired longitudinal slip, i.e.,
\begin{align}
    \frac{v_{f x}^{w}}{\omega_{f}(\Delta t / 2) r_{f}}-1=s_{f x}.\label{eq:slip_eqn}
\end{align}
Combining~\eqref{eq:lin_approx} and~\eqref{eq:slip_eqn}, we obtain the formula for determining the torque on the front wheel:
\begin{align}
	T_{f}=f_{f x}^{w} r_{f}+\frac{2 I_{f}}{\Delta t}\left(\frac{v_{f x}^{w}}{r_{f} (s_{f x}+1)}-\omega_{f}(0)\right),\label{eq:Tf}
\end{align}
where $\omega_f(0)$ refers to the current measured value of $\omega_f$. Completely similarly, for the rear wheel we can obtain
\begin{align}
	T_{r}=f_{r x}^{b} r_{r}+\frac{2 I_{r}}{\Delta t}\left(\frac{v_{r x}^{b}}{r_{r} (s_{r x}+1)}-\omega_{r}(0)\right).\label{eq:Tr}
\end{align}
So far, we have formulas for all the three input variables $\delta, T_f, T_r$ as specified in~\eqref{eq:delta}, \eqref{eq:Tf}, \eqref{eq:Tr}, which concludes the mechanism-based part of the controller design.

\subsection{Data-driven parameter estimator}

The controller relies on the tyre-road interaction parameters $\Theta = (B,C,D)$ in order to obtain precise control inputs as shown in~\eqref{eq:delta}, but those parameters are not known \textit{a priori}. Therefore, we adopt a data-driven approach to estimate the parameters.

Given a state-input pair $(\mathbf{X}, \mathbf{U})$, and the current estimate of parameters ${{\Theta}}$, we can predict $\hat{\mathbf{Y}}\left(\mathbf{X}, \mathbf{U}; {{\Theta}}\right) = \left[ \hat{\ddot{x}}, \hat{\ddot{y}}, \hat{\ddot{\psi}}, \hat{\dot{\omega}}_f,\hat{\dot{\omega}}_r \right]^\top$ according to~\eqref{eq:xddot}-\eqref{eq:omegadot}. Meanwhile, by applying the input $\mathbf{U}$ to the car at state $\mathbf{X}$, we can obtain the actual $\mathbf{Y} = \left[ \ddot{x}, \ddot{y}, \ddot{\psi}, \dot{\omega}_f, \dot{\omega}_r \right]^\top$. The data-driven module learns the parameters by collecting the trajectory data while running the controller and minimizing the error between the prediction and the actual value. Formally, let the collected dataset be $\boldsymbol{D}=\left\{\left(\mathbf{X}^{(i)}, \mathbf{U}^{(i)}\right), \mathbf{Y}^{(i)}\right\}_{i=1}^{n}$, then the parameter update rule can be specified as
\begin{align*}
    \underset{\Theta}{\operatorname{min}}\; \mathcal{L}(\Theta ; \mathcal{D}):=\sum_{i=1}^{n} \ell\left(\hat{\mathbf{Y}}\left(\mathbf{X}^{(i)}, \mathbf{U}^{(i)}; {{\Theta}}\right), \mathbf{Y}^{(i)}\right)+\mathcal{R}(\Theta),
\end{align*}
where $\ell\left( \hat{\mathbf{Y}}, \mathbf{Y} \right)$ is a loss function, for which we use Huber loss~\cite{huber1992robust} which is robust to modeling and measurement errors, and $\mathcal{R}$ is a regularization term to enforce sensibility of the parameters, for which we use logarithm barriers. Once a new batch of data is available, the parameter update can be performed by stepping in a nonlinear optimization algorithm, e.g. L-BFGS~\cite{Liu_Nocedal_1989}.

It shall be pointed out that $\Theta$ can also include parameters that describe other unmodelled effects, such as aerodynamics and suspension, and therefore the data-driven module can complement the mechanism-based module by making up for the simplicity of the model.

%% file: figures/framework.tikz
\usetikzlibrary{shapes,fit,positioning,backgrounds}

 \begin{tikzpicture}
    \node [draw=thupurple!50,fill=thupurple!20,thick,rounded corners, minimum width=3.5cm] (tyre) at (0,0) {Tyre characteristics};
    \node [draw=thupurple!50,fill=thupurple!20,thick,rounded corners, minimum width=3.5cm] (rigid) at (6,0) {Rigid-body dynamics};
    \node [draw=thupurple!50,fill=thupurple!20,thick,rounded corners, minimum width=3.5cm] (tyre_model) at (0,-2) {Approx. tyre model};
    \node [draw=thupurple!50,fill=thupurple!20,thick,rounded corners, minimum width=3.5cm] (mpc) at (6,-2) {MPC};
    \node [draw=thupurple!50,fill=thupurple!20,thick,rounded corners, minimum width=3.5cm] (learner) at (6,-4) {Param. update};
    \node [draw=thupurple!50,fill=thupurple!20,thick,rounded corners, minimum width=3.5cm] (planner) at (9,-6) {Path planner};
    \node [minimum width=3.5cm](state) at (10,0) {Pose, (angular) velocity};
    \node [minimum width=2.5cm](data) at (10,-4) {Training data};
    \draw [-stealth, semithick] (tyre)--node[above,xshift=0.5cm]{Forces} (rigid);
    \draw [-stealth, semithick] (rigid)--(state);
    \draw [-stealth, semithick] ($(state.south west)!0.5!(state.south)$)|-($(mpc.east)!0.5!(mpc.north east)$);
    \draw [-stealth, semithick] (mpc)--node[above,xshift=0.5cm,text width=1cm]{Target forces} (tyre_model);
    \draw[-stealth, semithick](tyre_model) -- (-3.5,-2) |- (-3.5, 0) -- node[above, text width=1cm]{Control input} (tyre.west);
    \draw [-stealth, semithick, dashed] (state.south)--(data.north);
    \draw [-stealth, semithick, dashed] (data)--(learner);
    \draw [-stealth, semithick, dashed] (learner)-|(tyre_model);
    \draw [-stealth, semithick, dashed] (learner)|-(planner);
    \draw [-stealth, semithick] ($(planner.north)!0.3!(planner.north west)$)|-($(mpc.east)!0.5!(mpc.south east)$);
    \node [draw=thupurple, dash dot, thick, rounded corners, minimum width=10cm, minimum height=1.5cm] (car) at (3, 0.25) {};
    \node[below right] at (car.north west) {Controlled car};
    \node [draw=thupurple, dash dot, thick, rounded corners, minimum width=13cm, minimum height=4cm] (controller) at (4.5, -3) {};
    \node[above right] at (controller.south west) {Controller};
      
      
      \node [minimum width=3cm](mech) at (-1,-5.5) {Mechanism-based};
      \node [minimum width=3cm](dd) at (-1,-6) {Data-driven};
      \draw [-stealth, semithick](-3,-5.5) -- (mech);
      \draw [-stealth,dashed, semithick](-3,-6) -- (dd);
      
      \draw [thupurple, thick, dash dot] (2.5, 2) -- (2.5, -6.5);
      \node at (0, 1.5) {Fast scale};
      \node at (6, 1.5) {Slow scale};

\end{tikzpicture}

%% file: figures/ill_traj.tikz
\begin{tikzpicture}[scale=0.8]

\definecolor{color0}{rgb}{0.12156862745098,0.466666666666667,0.705882352941177}
\definecolor{color1}{rgb}{1,0.498039215686275,0.0549019607843137}

\begin{axis}[
legend cell align={left},
legend style={fill opacity=0.8, draw opacity=1, text opacity=1, at={(0.03,0.97)}, anchor=north west, draw=white!80!black},
tick align=outside,
tick pos=left,
x grid style={white!69.0196078431373!black},
xlabel={$x(\mathrm{m})$},
xmin=-5.71328246705683, xmax=119.978931808193,
xtick style={color=black},
y grid style={white!69.0196078431373!black},
ylabel={$y(\mathrm{m})$},
ymin=-4.07476617750948, ymax=56.3502495801449,
ytick style={color=black}
]
\addplot [semithick, color0]
table {%
0 0
0 0
0.00262987360211777 -3.44508945441297e-28
0.0131466906863049 -3.50589264888241e-28
0.0368089669752906 -3.64269658630088e-28
0.0788742408667962 -3.88589783314139e-28
0.144599681177781 -4.26589081910953e-28
0.239242202702048 -4.81306850533105e-28
0.368058003207038 -5.55781970732678e-28
0.536303628760187 -6.53053525447717e-28
0.74923426860197 -7.76159813146158e-28
1.01210507188124 -9.28139109129176e-28
1.33017120143356 -1.11202969660625e-27
1.70868717423114 -1.33086948539516e-27
2.15290708246649 -1.58769613969148e-27
2.66808461664288 -1.88554709147185e-27
3.25945925733183 -2.22745155722037e-27
3.93231153791911 -2.61646248819264e-27
4.69187970987618 -3.05560839025567e-27
5.54341655088947 -3.54792616762603e-27
6.49217282940422 -4.09645156290033e-27
7.54339946251994 -4.70422040460016e-27
8.69977168055858 -5.37277938264438e-27
9.96124281111858 -6.10210151329367e-27
11.3278608189455 -6.89221452746275e-27
12.7996000534583 -7.7431035952109e-27
14.3764803271758 -8.6547801711899e-27
16.0584410583511 -9.62720922994045e-27
17.8455542052205 -1.06604323742592e-26
19.7377644878345 -1.17544176439343e-26
21.7352071805471 -1.29092432481084e-26
23.8376570404021 -1.41247789621088e-26
26.0400856326452 -1.5398117586026e-26
28.3319875211586 -1.67231853803474e-26
30.702811258606 -1.8093882004076e-26
33.1420814244273 -1.95041510712538e-26
35.6392907983532 -2.09479178108547e-26
38.1839250335644 -2.24191033316233e-26
40.7655585087965 -2.39116800391664e-26
43.3736016287599 -2.54195255371703e-26
45.997532679421 -2.69365566750094e-26
48.6268484405712 -2.84567009943009e-26
51.2561597807525 0.00468800961606795
53.8853860930914 0.0225688498212583
56.5142991722479 0.062097138821615
59.1424490829499 0.131713748673595
61.7690967023959 0.23989144335086
64.3930559732704 0.394993644905824
67.0128315353457 0.605513220730433
69.6260205423291 0.879582384201426
72.2299639836482 1.22530409958401
74.8211777748529 1.65064107159218
77.3954112351872 2.16334727310523
79.9475880915408 2.77085683646158
82.4717526768419 3.48021330122719
84.9610157470356 4.29799938503728
87.4075228134539 5.23020831422427
89.8024348643891 6.28211619441507
92.1359248544836 7.45814540817227
94.3971926011667 8.7617185100851
96.5745041342066 10.1950997991496
98.6552569827835 11.7592283761539
100.626073826123 13.4535483697512
102.47292770223 15.2758387942709
104.181296368106 17.2220518853025
105.736353935851 19.2861576087067
107.123198118993 21.4600046306848
108.353865597671 23.721382473297
109.448331776283 26.048462933553
110.411752747721 28.4299590583653
111.249558670792 30.8560662539393
111.967777868303 33.3181396987719
112.57297894252 35.8086211147428
113.072158730779 38.3209867686731
113.472641735366 40.8496894096149
113.78203152093 43.3900834229878
114.008140421054 45.9383640545466
114.158682816843 48.4916257079234
114.241712403468 51.0472170128268
114.265649341137 53.603657954797
};
\addlegendentry{Reference}
\addplot [semithick, color1]
table {%
0 0
0 0
0.00262987360211777 -3.44508945441297e-28
0.0131466906863049 -3.50589264888241e-28
0.0368089669752906 -3.64269658630088e-28
0.0788742408667962 -3.88589783314139e-28
0.144679182814606 -1.61115368817673e-05
0.237144468379024 -0.000787922980937678
0.355818943540575 0.00694354124600803
0.487866571091353 0.0164853261278912
0.66184679791527 0.030185390779815
0.88195644539491 0.047221843480635
1.11604636746215 0.0692721612558125
1.3654218279158 0.0758726437612772
1.61397377179626 0.121023582946557
1.86490375976696 0.184143644250088
2.17376391361125 0.225469806242937
2.58696906095357 0.27840292374024
3.09347578193244 0.360325401167944
3.64509082143481 0.4623449264213
4.21870383391912 0.580698366053427
4.83640050980083 0.773604528654738
5.50928169302194 1.07684830269039
6.20954616150911 1.4507242674183
6.90629686106787 1.86936559796897
7.60131936170152 2.33373947965224
8.32652415810894 2.74201624928554
9.08237561022708 3.11868025880422
9.81319934279626 3.46715895752515
10.4428452451604 3.7607089396447
10.9479951797548 3.99671192700913
11.4074943431204 4.15905158341592
11.9209159927763 4.30939436536832
12.4830154647828 4.45762813676672
13.0960855618054 4.5508316573565
13.7938371230002 4.59772479247929
14.5430639995837 4.60778462029583
15.3023245106005 4.6107983435206
16.0627096975866 4.59841270951939
16.7959880859818 4.48932228539907
17.4692380202098 4.34774294348187
18.0964348301784 4.1833469847658
18.737306240405 4.04200772974621
19.4280626641328 3.92872665255709
20.1791885974083 3.83819017102423
20.9716731912594 3.75866088054515
21.7996301801855 3.71806934630507
22.6447854805734 3.75302889458878
23.4992163926553 3.80483071668146
24.3979803549796 3.88845134670208
25.3551670183341 3.98539725261986
26.3317115940681 4.01252550096896
27.2900600820898 3.88920467978209
28.2432513851936 3.67791150398393
29.1983360284848 3.37758124593106
30.1047390100481 3.04222468675568
30.919761296339 2.62343791930368
31.7092017643794 2.1206972267887
32.4984487200393 1.58256622932938
33.3053179897521 1.10691599552272
34.2004694927374 0.732240164476566
35.1788135902674 0.394168995061402
36.150114332628 0.0238749433122816
37.1037596202957 -0.34593967612503
38.0848016590845 -0.640501941123678
39.031518839044 -0.934254399543711
39.9031520789017 -1.18227403076479
40.7366877669956 -1.29860158294281
41.6220252671051 -1.32765134918039
42.5555840004728 -1.32817455216156
43.4405617116061 -1.26662782323483
44.2747610862564 -1.15576682090429
45.1201729469618 -0.989019283009581
45.9600542237455 -0.700152487977737
46.7477990818361 -0.372729207418535
47.4917111353296 0.0131092480356727
48.2450502640252 0.492806951331229
48.9967512758453 1.0353942527087
49.746023274402 1.6341885637872
50.5003868036891 2.31184244766789
};
\addlegendentry{Actual}
\end{axis}

\end{tikzpicture}

%% file: figures/ill_time.tikz
\begin{tikzpicture}[scale=0.8]

\definecolor{color0}{rgb}{0.12156862745098,0.466666666666667,0.705882352941177}
\definecolor{color1}{rgb}{1,0.498039215686275,0.0549019607843137}

\begin{axis}[
tick align=outside,
tick pos=left,
x grid style={white!69.0196078431373!black},
xlabel={Simulation time (s)},
xmin=-0.8, xmax=16.8,
xtick style={color=black},
y grid style={white!69.0196078431373!black},
ylabel={Compution time (s)},
ymin=-6.89192359715089, ymax=144.73797106415,
ytick style={color=black}
]
\addplot [semithick, color0]
table {%
0 0.000344341999152675
0.20253164556962 2.12846755981445
0.405063291139241 2.12967824935913
0.607594936708861 2.13063454627991
0.810126582278481 2.13153600692749
1.0126582278481 4.91065883636475
1.21518987341772 5.28507995605469
1.41772151898734 5.92782020568848
1.62025316455696 7.05443239212036
1.82278481012658 9.16758632659912
2.0253164556962 11.8910846710205
2.22784810126582 14.9279479980469
2.43037974683544 15.660343170166
2.63291139240506 19.2469482421875
2.83544303797468 19.7612133026123
3.0379746835443 20.5812377929688
3.24050632911392 23.3277912139893
3.44303797468354 24.430004119873
3.64556962025316 24.8017635345459
3.84810126582278 28.1688976287842
4.05063291139241 28.5461826324463
4.25316455696202 29.9645252227783
4.45569620253165 31.434534072876
4.65822784810127 31.7581768035889
4.86075949367089 34.5863075256348
5.06329113924051 35.9378929138184
5.26582278481013 37.1152954101562
5.46835443037975 40.2636108398438
5.67088607594937 41.2638130187988
5.87341772151899 42.6383438110352
6.07594936708861 45.7951812744141
6.27848101265823 47.5555267333984
6.48101265822785 49.0132789611816
6.68354430379747 51.113151550293
6.88607594936709 53.585865020752
7.08860759493671 56.4161567687988
7.29113924050633 58.5600357055664
7.49367088607595 59.1416130065918
7.69620253164557 62.0853614807129
7.89873417721519 64.7211151123047
8.10126582278481 66.6730117797852
8.30379746835443 69.650764465332
8.50632911392405 73.1014938354492
8.70886075949367 73.7826766967773
8.91139240506329 76.2571411132812
9.11392405063291 77.9736099243164
9.31645569620253 78.5687789916992
9.51898734177215 78.8154678344727
9.72151898734177 81.8494110107422
9.92405063291139 82.9423141479492
10.126582278481 84.573112487793
10.3291139240506 84.7947845458984
10.5316455696203 87.4348983764648
10.7341772151899 87.8507995605469
10.9367088607595 90.4298477172852
11.1392405063291 93.0837249755859
11.3417721518987 95.7636108398438
11.5443037974684 97.8639602661133
11.746835443038 100.646385192871
11.9493670886076 103.293518066406
12.1518987341772 106.403579711914
12.3544303797468 107.297523498535
12.5569620253165 108.276947021484
12.7594936708861 109.727210998535
12.9620253164557 112.118064880371
13.1645569620253 113.675636291504
13.3670886075949 116.42219543457
13.5696202531646 118.060661315918
13.7721518987342 118.593170166016
13.9746835443038 121.347732543945
14.1772151898734 124.293632507324
14.379746835443 126.208763122559
14.5822784810127 127.119560241699
14.7848101265823 128.300827026367
14.9873417721519 129.547821044922
15.1898734177215 130.305969238281
15.3924050632911 133.070648193359
15.5949367088608 134.550338745117
15.7974683544304 135.221389770508
16 137.845703125
};
\end{axis}

\end{tikzpicture}

%% file: simulation.tex
\section{Simulation}

\subsection{Tracking task}

In this task, the goal is to drive a simulated car with the model described in Section~\ref{sec:problem} to track a manually specified reference trajectory. We use the geometric and mechanical parameters of a Mercedes CLS 63 AMG car and the tyre-road interaction parameters of a typical tarmac road ($B=10,C=1.9,D=1$). We choose the control interval to be $\Delta t = 0.1 \mathrm{s}$.  We use this task to benchmark aggressive driving because it requires the car to turn two right angles with radii of about $10\mathrm{m}$ at the speed of about $90 \mathrm{km/h}$, and such aggressive cornering can only be achieved with precise control of the slip.

\begin{figure}[!htbp]
    \centering
    \includegraphics[width=0.6\columnwidth]{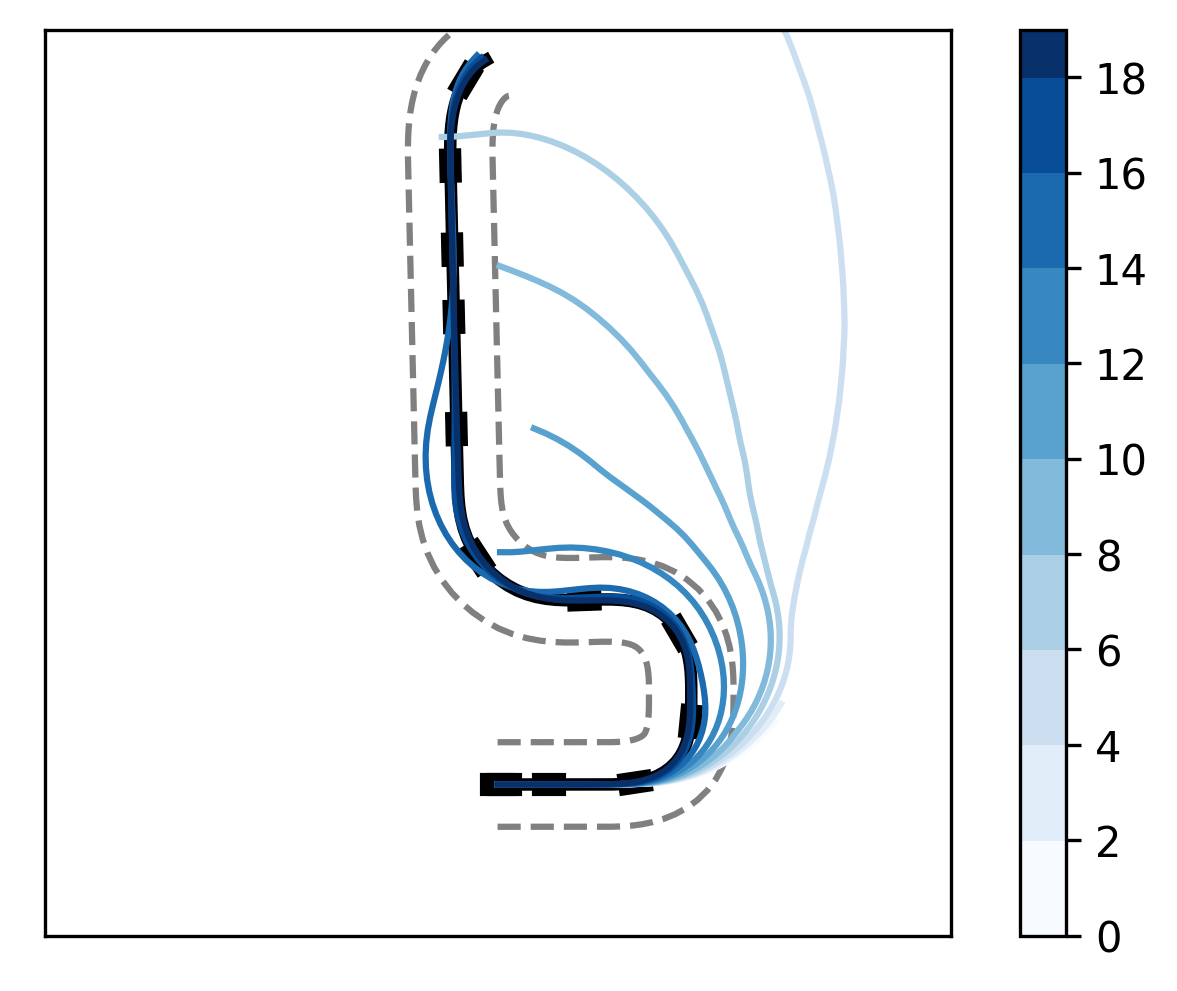}
    \caption{Iterative improvement of tracking performance of our proposed controller}
    \label{fig:tracking}
\end{figure}

The qualitative tracking performance of our proposed controller and its iterative improvement over trials is shown in Fig.~\ref{fig:tracking}. The gray dashed lines stand for the virtual lanes, the thick black line in the middle of the lane stands for the reference trajectory, and the blue lines, from light to dark, stand for the actual performance of different trials of our controller, from early to late. The numbers to the right of the color bar stand for the indices of corresponding trials. We can observe that the tracking performance improves steadily as the controller gets more trials, which proves the effectiveness of our data-driven strategy. We can also observe that the actual trajectory almost coincides with the reference trajectory after about 20 trials, which proves the correctness of our two-timescale mechanism-based controller design when the parameter estimates are accurate.

\begin{figure}[!htbp]
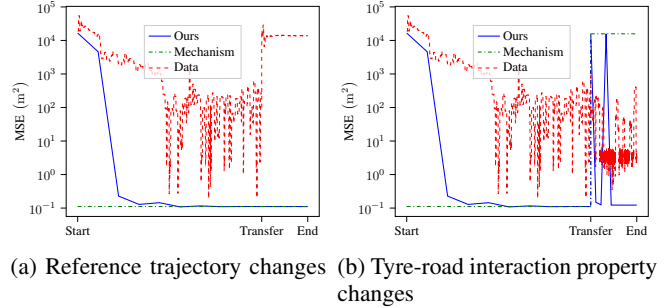

    \centering
    \subcaptionbox{Reference trajectory changes\label{fig:transfer1}}{\resizebox{0.48\columnwidth}{!}{\input{figures/transfer1.tikz}}}
    \subcaptionbox{Tyre-road interaction property changes\label{fig:transfer3}}{\resizebox{0.48\columnwidth}{!}{\input{figures/transfer3.tikz}}}
    \caption{Comparison of data efficiency and ability of transfer}
    \label{fig:transfer}
\end{figure}

Next we compare quantitative performance of our proposed method with purely mechanism-based or purely data-driven method, in terms of data efficiency and ability of transfer. We use our proposed mechanism-based controller with access to the true initial tyre-road interaction parameters $B,C,D$ as the mechanism-based baseline, and use a state-of-the-art deep reinforcement learning algorithm, SAC~\cite{Haarnoja_Zhou_Abbeel_Levine_2018}, as the data-driven baseline. In the experiment, we first allow each method 1M training samples in the process of repetitively attempting to complete the tracking task, after which we change some experiment conditions (marked by ``Transfer'' in Fig.~\ref{fig:transfer}), and allow each method another 1M training samples for adapting to the changes. The results of this experiment are presented in Fig.~\ref{fig:transfer}. For the ease of observation, both the horizontal and vertical axes in Fig.~\ref{fig:transfer} are in logarithm scale.

We can observe from Fig.~\ref{fig:transfer} that the tracking Mean-Square Error (MSE) converges to about $0.1 \mathrm{m}^2$ with significantly less training samples than the data-driven method, which demonstrates the benefit of our method in terms of data efficiency. In Fig.~\ref{fig:transfer1}, we test the ability of transfer by flipping the reference trajectory upside-down. Both the mechanism-based method and our proposed method are unaffected by this change, because they do not depend on a particular reference trajectory. By contrast, the MSE of the data-driven method become very large after this change, which can be explained by the fact that data-driven methods may overfit to a particular task unless they are exposed to data from sufficiently diverse tasks. This demonstrates the benefit of our method over purely data-driven methods in terms of generalization or ability of transfer. In Fig.~\ref{fig:transfer3}, we test the ability of transfer by changing the tyre-road interaction parameters $B,C,D$. The change can be viewed as significant because the purely mechanism-based method, which does not perform parameter update, almost completely fails to track the reference trajectory after the change. By contrast, our proposed method adapts to this change after performing parameter update with some additional training data, and the adaptation is significantly faster than that of the purely data-driven method. This also demonstrates the benefit of our method over both purely mechanism-based and purely data-driven methods in terms of adaptability.

\subsection{Racing task}

\begin{figure}[!htbp]
    \centering
    \subcaptionbox{Track \#1}{\resizebox{0.48\columnwidth}{!}{\input{figures/torcs1.tikz}}}
    \subcaptionbox{Track \#2}{\resizebox{0.48\columnwidth}{!}{\input{figures/torcs2.tikz}}}
    \subcaptionbox{Track \#3}{\resizebox{0.48\columnwidth}{!}{\input{figures/torcs3.tikz}}}
    \caption{Result on TORCS (1 sample=0.2s simulation time)}
    \label{fig:torcs}
\end{figure}
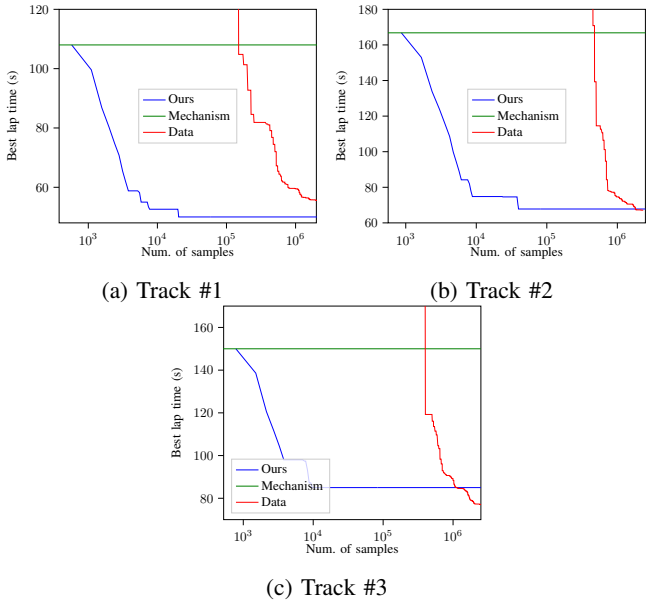

In this task, the goal is to pursue the minimum lap time in TORCS (The Open Racing Car Simulator) software, which uses a high-fidelity simulation model more complicated than the one described in Section~\ref{sec:problem}. We compare the lap times achieved by our proposed method on three different tracks with those achieved by purely mechanism-based and purely data-driven methods. We use our proposed mechanism-based controller as the mechanism-based baseline, and use a state-of-the-art deep reinforcement learning algorithm, SAC~\cite{Haarnoja_Zhou_Abbeel_Levine_2018}, as the data-driven baseline. The mechanism-based baseline uses $B,C,D$ values of a very slippery road, which is a design choice that errs on the side of caution and ensures the safety of the car. We record the best lap time over the entire training process. The results are presented in Fig.~\ref{fig:torcs}. We can observe that our proposed method can significantly improve over the mechanism-based method in terms of racing performance, and its final lap time is on par with that of the data-driven method. Furthermore, the data efficiency of our proposed method is about 100X that of the data-driven method. 

\begin{figure}[!htbp]
    \centering
    \includegraphics[width=\columnwidth]{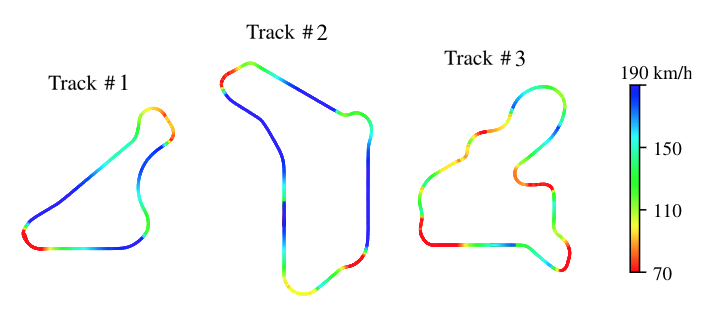}
    \caption{Speed profiles of our proposed controller on TORCS}
    \label{fig:speed}
\end{figure}

The speed profiles of our proposed controller after sufficient trials are shown in Fig.~\ref{fig:speed}. We can observe that the controller learns to accelerate on straight lanes and decelerate while cornering, which is sensible behavior from human perspective. A video demonstrating the learning process of our proposed and its comparison with that of the purely data-driven method is available at \url{https://www.bilibili.com/video/BV1L5411u79h}.

\section{Conclusion}

In this paper, we propose a method for the control of aggressive driving of autonomous cars, which features a modular design that is consisted of mechanism-based and data-driven components, and aware of the two-timescale phenomenon in the car model. Experiment results on a tracking task and a racing task show that our proposed method enjoys benefits over existing mechanism-based and data-driven methods in terms of data efficiency, ability of transfer and final performance.

%% file: figures/torcs1.tikz
\begin{tikzpicture}[scale=0.8]

\begin{axis}[
legend cell align={left},
legend style={fill opacity=0.8, draw opacity=1, text opacity=1, at={(0.5,0.5)}, anchor=center, draw=white!80!black},
log basis x={10},
tick align=outside,
tick pos=left,
x grid style={white!69.0196078431373!black},
xlabel={Num. of samples},
xmin=378.228487899011, xmax=2000000,
xmode=log,
xtick style={color=black},
xtick={10,100,1000,10000,100000,1000000,10000000,100000000},
xticklabels={\(\displaystyle {10^{1}}\),\(\displaystyle {10^{2}}\),\(\displaystyle {10^{3}}\),\(\displaystyle {10^{4}}\),\(\displaystyle {10^{5}}\),\(\displaystyle {10^{6}}\),\(\displaystyle {10^{7}}\),\(\displaystyle {10^{8}}\)},
y grid style={white!69.0196078431373!black},
ylabel={Best lap time (s)},
ymin=48, ymax=120,
ytick style={color=black}
]
\addplot [semithick, blue]
table {%
575 108
1106 99.6
1570 86.8
2001 80.2
2404 74.8
2787 70.8
3142 65.4
3479 61.8
3801 58.8
4862 58.8
5197 58.8
5516 58.2
5819 55
6827 55
7133 55
7428 53.4
7719 52.6
9148 52.6
10506 52.6
10832 52.6
11814 52.6
12144 52.6
12740 52.6
14783 52.6
15162 52.6
15522 52.6
15865 52.6
16196 52.6
16514 52.6
16824 52.6
18820 52.6
19147 52.6
19456 52.6
19751 52.6
20039 51.6
20318 50
21423 50
22456 50
22775 50
24671 50
25120 50
25569 50
25920 50
26912 50
27255 50
27583 50
27897 50
28204 50
28503 50
28796 50
29081 50
29362 50
30364 50
30663 50
30958 50
31239 50
31519 50
32852 50
33164 50
35209 50
35669 50
36026 50
36361 50
36684 50
38643 50
38978 50
39303 50
39609 50
39905 50
40358 50
40670 50
40975 50
42120 50
42444 50
43447 50
45325 50
45672 50
46003 50
46323 50
46632 50
46932 50
47224 50
47512 50
48066 50
49413 50
49745 50
50069 50
50378 50
50885 50
51202 50
51509 50
51807 50
52098 50
53101 50
53413 50
53718 50
54011 50
55998 50
56320 50
58258 50
};
\addlegendentry{Ours}
\addplot [semithick, blue, forget plot]
table {%
58258 50
2500000 50
};
\addplot [semithick, green!50!black]
table {%
1 108
2500000 108
};
\addlegendentry{Mechanism}
\addplot [semithick, red]
table {%
840 160
2375 160
2685 160
2935 160
3240 160
7185 160
8690 160
9390 160
11540 160
11950 160
12435 160
16235 160
22465 160
25675 160
29540 160
29955 160
33500 160
33560 160
34835 160
37320 160
38995 160
39170 160
40905 160
41110 160
42495 160
46305 160
47345 160
52000 160
54530 160
54835 160
57665 160
57835 160
59850 160
64390 160
65175 160
71475 160
73345 160
73900 160
77370 160
80585 160
84290 160
85490 160
86590 160
87720 160
97660 153.46875
98205 153.46875
98660 153.46875
100315 136.2171875
102700 136.2171875
102895 136.2171875
103915 136.2171875
104280 136.2171875
105405 136.2171875
105725 136.2171875
106600 136.2171875
111655 136.2171875
111795 136.2171875
116435 136.2171875
116810 136.2171875
117140 136.2171875
118355 136.2171875
118510 136.2171875
120740 136.2171875
121415 136.2171875
121615 136.2171875
123640 136.2171875
124675 136.2171875
126270 126.0828125
133990 126.0828125
138020 126.0828125
144855 126.0828125
144985 126.0828125
146320 126.0828125
149150 126.0828125
150030 104.8046875
152355 104.8046875
153065 104.8046875
153535 104.8046875
153605 104.8046875
157035 104.8046875
157220 104.8046875
159685 104.8046875
162180 104.8046875
164475 104.8046875
164515 104.8046875
165910 104.8046875
166880 104.8046875
168690 104.8046875
168750 104.8046875
168930 104.8046875
169200 104.8046875
169385 104.8046875
169570 104.8046875
171815 104.8046875
172630 104.8046875
173000 104.8046875
177420 101.328125
179770 101.328125
180190 101.328125
183150 101.328125
183300 101.328125
184255 101.328125
186015 101.328125
187450 101.328125
188825 101.328125
189385 101.328125
193605 101.328125
194205 101.328125
196485 101.328125
198515 101.328125
203605 92.734375
203970 92.734375
206360 92.734375
210675 92.734375
212345 92.734375
218200 92.734375
218560 92.734375
219030 92.734375
219970 92.734375
221005 92.734375
221260 92.734375
221525 92.734375
224085 92.734375
224825 92.734375
226865 84.5796875
227540 84.5796875
228295 84.5796875
231675 84.5796875
232070 84.5796875
236510 84.5796875
238725 84.5796875
239855 84.5796875
240910 84.5796875
246365 84.5796875
246760 84.5796875
250285 81.878125
254100 81.878125
255675 81.878125
256140 81.878125
258995 81.878125
263415 81.878125
264535 81.878125
264700 81.878125
265465 81.878125
267345 81.878125
269315 81.878125
274040 81.878125
278835 81.871875
281690 81.871875
282610 81.871875
284105 81.871875
293600 81.871875
293735 81.871875
293820 81.871875
296325 81.871875
300900 81.871875
300930 81.871875
302380 81.871875
310700 81.871875
312740 81.871875
318360 81.871875
322610 81.871875
324450 81.871875
328220 81.871875
333940 81.871875
334885 81.871875
335110 81.871875
337205 81.871875
345930 81.871875
347280 81.871875
348600 81.871875
349580 81.871875
352315 81.871875
352905 81.871875
353285 81.871875
354985 81.871875
357145 81.871875
359535 81.871875
359790 81.871875
359900 81.871875
360065 81.871875
360190 81.871875
361490 81.871875
363620 81.871875
365050 81.871875
377440 81.3390625
381095 81.3390625
382415 81.3390625
383370 81.3390625
384535 81.3390625
384570 81.3390625
391045 81.3390625
393880 81.3390625
394605 81.3390625
394870 81.3390625
395030 81.3390625
400505 81.04375
401955 81.04375
402035 81.04375
403070 81.04375
404985 81.04375
405080 81.04375
413855 81.04375
416340 81.04375
422050 81.04375
423010 81.04375
425450 79.1046875
430075 79.1046875
433905 79.1046875
434330 79.1046875
435045 79.1046875
436970 79.1046875
440690 79.1046875
443375 79.1046875
446740 79.1046875
446820 79.1046875
449870 79.1046875
451190 76.8171875
456505 76.8171875
458125 76.8171875
458240 76.8171875
460605 76.8171875
463945 76.8171875
466120 76.8171875
467360 76.8171875
469510 76.8171875
473925 76.8171875
475290 74.4734375
476890 74.4734375
477405 74.4734375
481510 74.4734375
482680 74.4734375
488985 74.4734375
489935 74.4734375
491065 74.4734375
492375 74.4734375
495280 74.4734375
499425 74.4734375
502950 72.0140625
504295 72.0140625
504590 72.0140625
504990 72.0140625
505365 72.0140625
506680 72.0140625
509620 72.0140625
509720 72.0140625
514405 72.0140625
514560 72.0140625
514835 72.0140625
515010 72.0140625
515060 72.0140625
517590 72.0140625
519740 72.0140625
521385 72.0140625
521675 72.0140625
524485 72.0140625
524625 72.0140625
525470 67.228125
527175 67.228125
528875 67.228125
528945 67.228125
531340 67.228125
532835 67.228125
539190 67.228125
542935 67.228125
546510 67.228125
549115 67.228125
551125 65.4078125
552095 65.4078125
553210 65.4078125
556620 65.4078125
558495 65.4078125
564515 65.4078125
566615 65.4078125
568710 65.4078125
569905 65.4078125
571465 65.4078125
575485 64.4234375
577535 64.4234375
577595 64.4234375
579395 64.4234375
583370 64.4234375
583560 64.4234375
583920 64.4234375
585240 64.4234375
588180 64.4234375
589130 64.4234375
589710 64.4234375
590955 64.4234375
594995 64.4234375
596515 64.4234375
605965 63.759375
605970 63.759375
606665 63.759375
607020 63.759375
609180 63.759375
611030 63.759375
613115 63.759375
613175 63.759375
615345 63.759375
618335 63.759375
620990 63.759375
626455 62.146875
637715 62.146875
638190 62.146875
645735 62.146875
646595 62.146875
647035 62.146875
647515 62.146875
650360 61.7609375
658405 61.7609375
658675 61.7609375
659385 61.7609375
663385 61.7609375
664585 61.7609375
669305 61.7609375
670040 61.7609375
671885 61.7609375
671950 61.7609375
672375 61.7609375
674390 61.7609375
678080 61.725
679365 61.725
681050 61.725
683370 61.725
683395 61.725
685280 61.725
687830 61.725
694700 61.725
696295 61.725
697020 61.725
697250 61.725
699880 61.725
700080 61.1703125
700230 61.1703125
700280 61.1703125
700660 61.1703125
703755 61.1703125
704950 61.1703125
706115 61.1703125
709450 61.1703125
710985 61.1703125
710990 61.1703125
713760 61.1703125
714465 61.1703125
715935 61.1703125
716715 61.1703125
718965 61.1703125
719595 61.1703125
721135 61.1703125
721430 61.1703125
722025 61.1703125
731565 61.0125
733485 61.0125
737460 61.0125
740950 61.0125
742735 61.0125
743025 61.0125
743370 61.0125
743390 61.0125
747215 61.0125
748420 61.0125
749435 61.0125
753795 60.9375
754320 60.9375
755215 60.9375
755285 60.9375
755465 60.9375
755670 60.9375
756140 60.9375
756385 60.9375
757720 60.9375
762620 60.9375
763835 60.9375
773325 60.9375
775950 59.9484375
777925 59.9484375
778605 59.9484375
778680 59.9484375
783265 59.9484375
785230 59.9484375
785325 59.9484375
785655 59.9484375
786460 59.9484375
787155 59.9484375
788120 59.9484375
797045 59.9484375
798900 59.9484375
801015 59.659375
801265 59.659375
802125 59.659375
802310 59.659375
803545 59.659375
805800 59.659375
805940 59.659375
806825 59.659375
809755 59.659375
810715 59.659375
811035 59.659375
812920 59.659375
813950 59.659375
816485 59.659375
823115 59.659375
827320 59.6328125
827990 59.6328125
829005 59.6328125
831495 59.6328125
831610 59.6328125
834740 59.6328125
836175 59.6328125
836985 59.6328125
838225 59.6328125
840190 59.6328125
840905 59.6328125
841950 59.6328125
843605 59.6328125
846620 59.6328125
847775 59.6328125
849370 59.6328125
849820 59.6328125
850945 59.63125
852480 59.63125
856065 59.63125
858105 59.63125
858300 59.63125
858350 59.63125
862085 59.63125
864865 59.63125
865470 59.63125
865875 59.63125
866480 59.63125
867785 59.63125
868730 59.63125
872075 59.63125
873240 59.63125
876370 59.63125
880410 59.63125
881420 59.63125
882920 59.63125
888430 59.63125
888890 59.63125
890965 59.63125
893310 59.63125
893510 59.63125
895985 59.63125
899835 59.63125
901580 59.63125
905250 59.63125
905270 59.63125
905545 59.63125
907320 59.63125
908185 59.63125
910440 59.63125
914180 59.63125
914450 59.63125
914725 59.63125
914750 59.63125
916125 59.63125
916360 59.63125
916370 59.63125
917435 59.63125
917995 59.63125
920750 59.63125
923730 59.63125
925295 59.63125
930030 59.63125
931165 59.63125
933445 59.63125
933585 59.63125
935455 59.63125
935705 59.63125
938005 59.63125
938375 59.63125
939720 59.63125
939950 59.63125
946265 59.63125
946300 59.63125
946615 59.63125
949670 59.63125
951695 59.63125
952215 59.63125
961080 59.63125
973110 59.63125
976490 59.559375
977005 59.559375
978100 59.559375
982485 59.559375
983635 59.559375
983885 59.559375
991010 59.559375
994260 59.559375
994280 59.559375
996300 59.559375
996440 59.559375
998315 59.559375
1001060 59.4890625
1001335 59.4890625
1007010 59.4890625
1008930 59.4890625
1009605 59.4890625
1011720 59.4890625
1012315 59.4890625
1014850 59.4890625
1016160 59.4890625
1023770 59.4890625
1027165 59.4890625
1030555 59.4890625
1035255 59.4890625
1038050 59.4890625
1041680 59.4890625
1043745 59.4890625
1044675 59.4890625
1047065 59.4890625
1049945 59.4890625
1050960 59.44375
1056210 59.44375
1058485 59.44375
1065500 59.44375
1065885 59.44375
1069340 59.44375
1070430 59.44375
1070605 59.44375
1073055 59.44375
1073640 59.44375
1076655 59.29375
1079770 59.29375
1080120 59.29375
1080265 59.29375
1087265 59.29375
1088510 59.29375
1089340 59.29375
1091105 59.29375
1091220 59.29375
1097575 59.29375
1098955 59.29375
1099310 59.29375
1101330 58.9734375
1103610 58.9734375
1104190 58.9734375
1106355 58.9734375
1109745 58.9734375
1110225 58.9734375
1117690 58.9734375
1124230 58.9734375
1126025 58.278125
1131240 58.278125
1131730 58.278125
1131790 58.278125
1132980 58.278125
1137060 58.278125
1139830 58.278125
1139840 58.278125
1140860 58.278125
1142920 58.278125
1142935 58.278125
1145000 58.278125
1146455 58.278125
1147225 58.278125
1147870 58.278125
1149690 58.278125
1151220 57.659375
1151860 57.659375
1155120 57.659375
1160305 57.659375
1160875 57.659375
1160965 57.659375
1162180 57.659375
1167355 57.659375
1167365 57.659375
1167430 57.659375
1168165 57.659375
1169050 57.659375
1169690 57.659375
1172295 57.659375
1174590 57.659375
1175300 57.6203125
1176080 57.6203125
1176550 57.6203125
1183695 57.6203125
1184535 57.6203125
1190240 57.6203125
1191530 57.6203125
1193080 57.6203125
1194075 57.6203125
1196695 57.6203125
1200475 57.1390625
1207660 57.1390625
1208150 57.1390625
1210325 57.1390625
1211710 57.1390625
1217860 57.1390625
1224505 57.1390625
1228575 56.6578125
1230255 56.6578125
1234700 56.6578125
1237540 56.6578125
1238175 56.6578125
1238475 56.6578125
1241465 56.6578125
1242180 56.6578125
1244805 56.6578125
1245200 56.6578125
1248385 56.6578125
1250125 56.6546875
1250225 56.6546875
1251215 56.6546875
1254350 56.6546875
1255980 56.6546875
1256600 56.6546875
1258480 56.6546875
1258985 56.6546875
1259050 56.6546875
1260335 56.6546875
1261700 56.6546875
1262930 56.6546875
1264820 56.6546875
1266005 56.6546875
1267200 56.6546875
1271475 56.6546875
1272115 56.6546875
1273010 56.6546875
1273455 56.6546875
1277315 56.6296875
1279980 56.6296875
1283655 56.6296875
1284730 56.6296875
1290315 56.6296875
1291040 56.6296875
1291755 56.6296875
1292035 56.6296875
1292845 56.6296875
1294800 56.6296875
1295560 56.6296875
1295615 56.6296875
1298685 56.6296875
1300070 56.59375
1300180 56.59375
1301060 56.59375
1302660 56.59375
1304770 56.59375
1311630 56.59375
1312565 56.59375
1312640 56.59375
1312995 56.59375
1314270 56.59375
1314320 56.59375
1316505 56.59375
1319740 56.59375
1321065 56.59375
1324805 56.59375
1326365 56.59375
1328010 56.59375
1328475 56.59375
1330625 56.59375
1331550 56.59375
1333295 56.59375
1335805 56.59375
1338220 56.59375
1338250 56.59375
1339035 56.59375
1339045 56.59375
1339490 56.59375
1343440 56.59375
1343585 56.59375
1346365 56.59375
1351710 56.59375
1352970 56.59375
1354310 56.59375
1355265 56.59375
1355930 56.59375
1357785 56.59375
1360370 56.59375
1361725 56.59375
1363760 56.59375
1364695 56.59375
1368920 56.59375
1377015 56.5921875
1377030 56.5921875
1377635 56.5921875
1379495 56.5921875
1383840 56.5921875
1387200 56.5921875
1393740 56.5921875
1394400 56.5921875
1395245 56.5921875
1397365 56.5921875
1397410 56.5921875
1397510 56.5921875
1399455 56.5921875
1402270 56.425
1403095 56.425
1404525 56.425
1406390 56.425
1407685 56.425
1408515 56.425
1411150 56.425
1413255 56.425
1418535 56.425
1421520 56.425
1424435 56.425
1427690 56.415625
1431060 56.415625
1431175 56.415625
1433520 56.415625
1435680 56.415625
1435720 56.415625
1436935 56.415625
1443805 56.415625
1444690 56.415625
1444875 56.415625
1446265 56.415625
1446825 56.415625
1449275 56.415625
1449930 56.415625
1454085 56.415625
1463800 56.415625
1465090 56.415625
1469135 56.415625
1470010 56.415625
1477045 56.415625
1477870 56.415625
1484410 56.415625
1485930 56.415625
1486500 56.415625
1491760 56.415625
1492070 56.415625
1495845 56.415625
1497430 56.415625
1497745 56.415625
1503575 56.415625
1504170 56.415625
1504970 56.415625
1505145 56.415625
1506295 56.415625
1509150 56.415625
1512160 56.415625
1512475 56.415625
1515335 56.415625
1520120 56.415625
1521400 56.415625
1522935 56.415625
1526525 56.3609375
1528800 56.3609375
1532535 56.3609375
1536355 56.3609375
1540625 56.3609375
1542060 56.3609375
1544280 56.3609375
1545050 56.3609375
1545340 56.3609375
1548740 56.3609375
1548825 56.3609375
1549585 56.3609375
1549725 56.3609375
1550535 56.10625
1551475 56.10625
1558200 56.10625
1560880 56.10625
1562940 56.10625
1563780 56.10625
1566690 56.10625
1568685 56.10625
1568965 56.10625
1569535 56.10625
1569740 56.10625
1571460 56.10625
1573905 56.10625
1574120 56.10625
1575850 56.053125
1577950 56.053125
1580985 56.053125
1583980 56.053125
1588490 56.053125
1588535 56.053125
1590035 56.053125
1592050 56.053125
1594145 56.053125
1595560 56.053125
1597180 56.053125
1599710 56.053125
1600070 55.875
1601415 55.875
1601960 55.875
1603530 55.875
1603720 55.875
1604295 55.875
1604640 55.875
1604780 55.875
1606920 55.875
1611945 55.875
1614645 55.875
1615045 55.875
1615250 55.875
1617640 55.875
1621525 55.875
1623450 55.875
1625315 55.7921875
1625510 55.7921875
1631900 55.7921875
1633385 55.7921875
1640195 55.7921875
1642720 55.7921875
1642725 55.7921875
1645255 55.7921875
1647715 55.7921875
1648555 55.7921875
1650240 55.7921875
1650565 55.7921875
1651660 55.7921875
1654850 55.7921875
1658195 55.7921875
1659125 55.7921875
1659750 55.7921875
1659885 55.7921875
1660640 55.7921875
1661295 55.7921875
1663605 55.7921875
1664035 55.7921875
1666290 55.7921875
1666630 55.7921875
1668820 55.7921875
1670500 55.7921875
1672215 55.7921875
1672245 55.7921875
1672365 55.7921875
1672390 55.7921875
1676780 55.7921875
1678135 55.7921875
1681920 55.7921875
1681965 55.7921875
1682135 55.7921875
1683625 55.7921875
1688225 55.7921875
1693545 55.7921875
1693685 55.7921875
1694455 55.7921875
1697730 55.7921875
1703580 55.790625
1708810 55.790625
1709040 55.790625
1709105 55.790625
1715330 55.790625
1715750 55.790625
1722520 55.790625
1722670 55.790625
1723610 55.790625
1725235 55.7734375
1726400 55.7734375
1726965 55.7734375
1735560 55.7734375
1740925 55.7734375
1741225 55.7734375
1742710 55.7734375
1744625 55.7734375
1746775 55.7734375
1760310 55.7734375
1762805 55.7734375
1763145 55.7734375
1763170 55.7734375
1766030 55.7734375
1771260 55.7734375
1783475 55.7734375
1789625 55.7734375
1790420 55.7734375
1790815 55.7734375
1790985 55.7734375
1792915 55.7734375
1794040 55.7734375
1794925 55.7734375
1795420 55.7734375
1796915 55.7734375
1797860 55.7734375
1798955 55.7734375
1801670 55.7734375
1802250 55.7734375
1803875 55.7734375
1805175 55.7734375
1805610 55.7734375
1809230 55.7734375
1809645 55.7734375
1811795 55.7734375
1813745 55.7734375
1815310 55.7734375
1819210 55.7734375
1819255 55.7734375
1821675 55.7734375
1823000 55.7734375
1823855 55.7734375
1823995 55.7734375
1826840 55.7734375
1833280 55.7734375
1834345 55.7734375
1841585 55.7734375
1843610 55.7734375
1845470 55.7734375
1848530 55.7734375
1850010 55.7578125
1850075 55.7578125
1854190 55.7578125
1855630 55.7578125
1856265 55.7578125
1856580 55.7578125
1858640 55.7578125
1869140 55.7578125
1871940 55.7578125
1874280 55.7578125
1875260 55.7515625
1876290 55.7515625
1877750 55.7515625
1879145 55.7515625
1882440 55.7515625
1884830 55.7515625
1885560 55.7515625
1885680 55.7515625
1886415 55.7515625
1886860 55.7515625
1889630 55.7515625
1891160 55.7515625
1896435 55.7515625
1898675 55.7515625
1899145 55.7515625
1900935 55.7515625
1901910 55.7515625
1903240 55.7515625
1908265 55.7515625
1911045 55.7515625
1913180 55.7515625
1914560 55.7515625
1915605 55.7515625
1919130 55.7515625
1919185 55.7515625
1922130 55.7515625
1922495 55.7515625
1924055 55.7515625
1924620 55.7515625
1927540 55.7515625
1928775 55.7515625
1930600 55.7515625
1931360 55.7515625
1932285 55.7515625
1937500 55.7515625
1938235 55.7515625
1938550 55.7515625
1938640 55.7515625
1939915 55.7515625
1940485 55.7515625
1942460 55.7515625
1944195 55.7515625
1944220 55.7515625
1945875 55.7515625
1948080 55.7515625
1956110 55.6015625
1961220 55.6015625
1965965 55.6015625
1967040 55.6015625
1971795 55.6015625
1974570 55.6015625
1975870 55.5734375
1978035 55.5734375
1980205 55.5734375
1980795 55.5734375
1984300 55.5734375
1987555 55.5734375
1988855 55.5734375
1989680 55.5734375
1990985 55.5734375
1991035 55.5734375
1996415 55.5734375
2000855 55.5734375
2000995 55.5734375
2002515 55.5734375
2003175 55.5734375
2004320 55.5734375
2006590 55.5734375
2008195 55.5734375
2008430 55.5734375
2015555 55.5734375
2016460 55.5734375
2017045 55.5734375
2017430 55.5734375
2021010 55.5734375
2023160 55.5734375
2023895 55.5734375
2025065 55.5734375
2026090 55.5734375
2028800 55.5734375
2029240 55.5734375
2032475 55.5734375
2036330 55.5734375
2037590 55.5734375
2037980 55.5734375
2038630 55.5734375
2039750 55.5734375
2042755 55.5734375
2046190 55.5734375
2049850 55.5734375
2055910 55.5609375
2056290 55.5609375
2060385 55.5609375
2061015 55.5609375
};
\addlegendentry{Data}
\end{axis}

\end{tikzpicture}

%% file: figures/torcs2.tikz
\begin{tikzpicture}[scale=0.8]

\begin{axis}[
legend cell align={left},
legend style={fill opacity=0.8, draw opacity=1, text opacity=1, at={(0.5,0.5)}, anchor=center, draw=white!80!black},
log basis x={10},
tick align=outside,
tick pos=left,
x grid style={white!69.0196078431373!black},
xlabel={Num. of samples},
xmin=563.113902928597, xmax=2500000,
xmode=log,
xtick style={color=black},
xtick={10,100,1000,10000,100000,1000000,10000000,100000000},
xticklabels={\(\displaystyle {10^{1}}\),\(\displaystyle {10^{2}}\),\(\displaystyle {10^{3}}\),\(\displaystyle {10^{4}}\),\(\displaystyle {10^{5}}\),\(\displaystyle {10^{6}}\),\(\displaystyle {10^{7}}\),\(\displaystyle {10^{8}}\)},
y grid style={white!69.0196078431373!black},
ylabel={Best lap time (s)},
ymin=60, ymax=180,
ytick style={color=black}
]
\addplot [semithick, blue]
table {%
867 166.8
1664 153
2363 133.8
3011 123.8
3617 115.4
4188 108.6
4715 99.8
5214 94.2
5688 89.2
6137 84.2
7097 84.2
7554 84.2
7992 82
8409 77.8
8811 74.8
10316 74.8
10738 74.8
11559 74.8
11980 74.8
12385 74.8
15113 74.8
15544 74.8
16314 74.8
16743 74.8
17156 74.8
17559 74.8
20265 74.8
22958 74.8
23366 74.8
23767 74.6
24539 74.6
24950 74.6
27358 74.6
27803 74.6
28233 74.6
28651 74.6
30157 74.6
30597 74.6
32970 74.6
33415 74.6
34207 74.6
34655 74.6
35387 74.6
35823 74.6
36571 74.6
36995 74.6
37415 74.6
37818 74.6
38211 73
38597 71.6
38976 70.2
39350 69.2
39717 67.8
40084 67.8
42801 67.8
43201 67.8
43933 67.8
44672 67.8
46969 67.8
47406 67.8
48126 67.8
49628 67.8
50098 67.8
50544 67.8
50978 67.8
51401 67.8
54073 67.8
54500 67.8
54919 67.8
55329 67.8
55725 67.8
56117 67.8
56502 67.8
57463 67.8
58424 67.8
59376 67.8
60917 67.8
61336 67.8
61740 67.8
62134 67.8
62928 67.8
63335 67.8
63737 67.8
65215 67.8
65618 67.8
66015 67.8
66403 67.8
66789 67.8
69511 67.8
72196 67.8
72605 67.8
73370 67.8
73791 67.8
76126 67.8
76567 67.8
77306 67.8
77749 67.8
80046 67.8
80483 67.8
80906 67.8
};
\addlegendentry{Ours}
\addplot [semithick, blue, forget plot]
table {%
80906 67.8
2500000 67.8
};
\addplot [semithick, green!50!black]
table {%
1 166.8
2500000 166.8
};
\addlegendentry{Mechanism}
\addplot [semithick, red]
table {%
840 200
2375 200
2685 200
2935 200
3240 200
7185 200
8690 200
11540 200
11950 200
12435 200
16235 200
22465 200
25675 200
29540 200
33500 200
33560 200
34835 200
37320 200
38995 200
39170 200
40905 200
41110 200
42495 200
46305 200
47345 200
52000 198.3390625
54530 198.3390625
54835 198.3390625
57665 198.3390625
59850 198.3390625
64390 198.3390625
65175 198.3390625
71475 198.3390625
73345 198.3390625
73900 198.3390625
80585 198.3390625
84290 198.3390625
86590 198.3390625
87720 198.3390625
97660 198.3390625
98205 198.3390625
98660 198.3390625
100315 198.3390625
102700 198.3390625
102895 198.3390625
104280 198.3390625
105405 198.3390625
105725 198.3390625
106600 198.3390625
111655 198.3390625
111795 198.3390625
116435 198.3390625
116810 198.3390625
117140 198.3390625
118355 198.3390625
118510 198.3390625
120740 198.3390625
121615 198.3390625
123640 198.3390625
124675 198.3390625
126270 198.3390625
133990 198.3390625
144855 198.3390625
144985 198.3390625
149150 198.3390625
150030 198.3390625
152355 198.3390625
153065 198.3390625
153535 198.3390625
157035 198.3390625
157220 198.3390625
159685 198.3390625
162180 198.3390625
164475 198.3390625
164515 198.3390625
165910 198.3390625
168690 198.3390625
168750 198.3390625
168930 198.3390625
169200 198.3390625
169385 198.3390625
169570 198.3390625
171815 198.3390625
172630 198.3390625
173000 198.3390625
177420 198.3390625
179770 198.3390625
180190 198.3390625
183150 198.3390625
183300 198.3390625
184255 198.3390625
186015 198.3390625
187450 198.3390625
188825 198.3390625
189385 198.3390625
193605 198.3390625
194205 198.3390625
196485 198.3390625
198515 198.3390625
203970 198.3390625
206360 198.3390625
210675 198.3390625
212345 198.3390625
218200 198.3390625
218560 198.3390625
219030 198.3390625
219970 198.3390625
221005 198.3390625
221260 198.3390625
221525 198.3390625
224085 198.3390625
224825 198.3390625
226865 198.3390625
227540 198.3390625
228295 198.3390625
231675 198.3390625
232070 198.3390625
238725 198.3390625
239855 198.3390625
240910 198.3390625
246365 198.3390625
250285 198.3390625
254100 198.3390625
255675 198.3390625
256140 198.3390625
258995 198.3390625
264535 198.3390625
264700 198.3390625
265465 198.3390625
267345 198.3390625
269315 198.3390625
274040 198.3390625
281690 198.3390625
282610 198.3390625
284105 198.3390625
293600 198.3390625
293735 198.3390625
293820 198.3390625
300900 198.3390625
300930 198.3390625
302380 198.3390625
310700 198.3390625
312740 198.3390625
318360 198.3390625
322610 198.3390625
328220 198.3390625
333940 198.3390625
335110 198.3390625
337205 198.3390625
345930 198.3390625
347280 198.3390625
348600 198.3390625
349580 198.3390625
352315 198.3390625
352905 198.3390625
353285 198.3390625
354985 198.3390625
357145 198.3390625
359535 198.3390625
359790 198.3390625
359900 198.3390625
360065 198.3390625
360190 198.3390625
363620 198.3390625
365050 198.3390625
377440 198.3390625
381095 198.3390625
383370 198.3390625
384535 198.3390625
384570 198.3390625
391045 198.3390625
394605 198.3390625
394870 198.3390625
395030 198.3390625
401955 198.3390625
403070 198.3390625
404985 198.3390625
405080 198.3390625
416340 198.3390625
422050 198.3390625
423010 198.3390625
425450 195.140625
430075 195.140625
433905 195.140625
434330 195.140625
435045 195.140625
436970 195.140625
440690 195.140625
443375 195.140625
446740 195.140625
446820 195.140625
449870 195.140625
451190 170.8453125
456505 170.8453125
458125 170.8453125
458240 170.8453125
460605 170.8453125
466120 170.8453125
467360 170.8453125
469510 170.8453125
473925 170.8453125
475290 139.3109375
476890 139.3109375
477405 139.3109375
481510 139.3109375
482680 139.3109375
489935 139.3109375
492375 139.3109375
495280 139.3109375
499425 139.3109375
502950 114.546875
504295 114.546875
504590 114.546875
505365 114.546875
506680 114.546875
509620 114.546875
509720 114.546875
514405 114.546875
514560 114.546875
514835 114.546875
515010 114.546875
515060 114.546875
519740 114.546875
521385 114.546875
521675 114.546875
524485 114.546875
524625 114.546875
525470 114.546875
527175 114.546875
528875 114.546875
531340 114.546875
532835 114.546875
539190 114.546875
542935 114.546875
549115 114.546875
551125 114.546875
552095 114.546875
553210 114.546875
556620 114.546875
558495 114.546875
564515 114.546875
566615 114.546875
568710 114.546875
569905 114.546875
571465 114.546875
575485 112.6484375
577535 112.6484375
577595 112.6484375
579395 112.6484375
583370 112.6484375
583560 112.6484375
583920 112.6484375
585240 112.6484375
588180 112.6484375
589130 112.6484375
589710 112.6484375
590955 112.6484375
605965 110.634375
605970 110.634375
606665 110.634375
607020 110.634375
609180 110.634375
611030 110.634375
613115 110.634375
613175 110.634375
615345 110.634375
618335 110.634375
620990 110.634375
626455 106.41875
637715 106.41875
638190 106.41875
645735 106.41875
646595 106.41875
647035 106.41875
647515 106.41875
650360 101.1921875
658405 101.1921875
658675 101.1921875
659385 101.1921875
669305 101.1921875
671885 101.1921875
671950 101.1921875
672375 101.1921875
674390 101.1921875
678080 94.6703125
681050 94.6703125
683370 94.6703125
683395 94.6703125
685280 94.6703125
687830 94.6703125
694700 94.6703125
696295 94.6703125
697020 94.6703125
697250 94.6703125
699880 94.6703125
700080 84.278125
700230 84.278125
700280 84.278125
700660 84.278125
703755 84.278125
704950 84.278125
706115 84.278125
709450 84.278125
710985 84.278125
710990 84.278125
713760 84.278125
714465 84.278125
715935 84.278125
719595 84.278125
721135 84.278125
721430 84.278125
722025 84.278125
731565 78.1921875
733485 78.1921875
737460 78.1921875
740950 78.1921875
742735 78.1921875
743025 78.1921875
743370 78.1921875
743390 78.1921875
747215 78.1921875
748420 78.1921875
749435 78.1921875
753795 77.9265625
754320 77.9265625
755215 77.9265625
755285 77.9265625
755465 77.9265625
755670 77.9265625
756140 77.9265625
756385 77.9265625
757720 77.9265625
762620 77.9265625
763835 77.9265625
775950 77.75625
777925 77.75625
778605 77.75625
778680 77.75625
783265 77.75625
785325 77.75625
785655 77.75625
786460 77.75625
787155 77.75625
788120 77.75625
797045 77.75625
798900 77.75625
801015 77.5125
801265 77.5125
802125 77.5125
803545 77.5125
805800 77.5125
805940 77.5125
806825 77.5125
809755 77.5125
811035 77.5125
812920 77.5125
813950 77.5125
816485 77.5125
823115 77.5125
827320 77.253125
827990 77.253125
829005 77.253125
831495 77.253125
831610 77.253125
834740 77.253125
836175 77.253125
836985 77.253125
838225 77.253125
840190 77.253125
840905 77.253125
843605 77.253125
846620 77.253125
847775 77.253125
849370 77.253125
849820 77.253125
850945 77.221875
852480 77.221875
856065 77.221875
858105 77.221875
858300 77.221875
858350 77.221875
864865 77.221875
865470 77.221875
865875 77.221875
867785 77.221875
868730 77.221875
872075 77.221875
873240 77.221875
880410 77.221875
881420 77.221875
882920 77.221875
888890 77.221875
890965 77.221875
893310 77.221875
893510 77.221875
895985 77.221875
899835 77.221875
901580 76.734375
905250 76.734375
905545 76.734375
907320 76.734375
908185 76.734375
910440 76.734375
914180 76.734375
914725 76.734375
914750 76.734375
916125 76.734375
916360 76.734375
916370 76.734375
917435 76.734375
917995 76.734375
923730 76.734375
925295 75.403125
930030 75.403125
931165 75.403125
933445 75.403125
935455 75.403125
935705 75.403125
938005 75.403125
938375 75.403125
939720 75.403125
939950 75.403125
946265 75.403125
946300 75.403125
946615 75.403125
949670 75.403125
951695 74.9375
952215 74.9375
961080 74.9375
973110 74.9375
976490 74.6140625
977005 74.6140625
978100 74.6140625
982485 74.6140625
983635 74.6140625
983885 74.6140625
991010 74.6140625
994260 74.6140625
994280 74.6140625
996440 74.6140625
998315 74.6140625
1001060 74.6140625
1007010 74.6140625
1008930 74.6140625
1009605 74.6140625
1014850 74.6140625
1023770 74.6140625
1027165 74.6140625
1030555 74.6140625
1035255 74.6140625
1038050 74.6140625
1041680 74.5234375
1043745 74.5234375
1044675 74.5234375
1049945 74.5234375
1050960 74.296875
1056210 74.296875
1058485 74.296875
1065500 74.296875
1065885 74.296875
1069340 74.296875
1070430 74.296875
1070605 74.296875
1073055 74.296875
1073640 74.296875
1076655 73.640625
1079770 73.640625
1080120 73.640625
1080265 73.640625
1089340 73.640625
1091105 73.640625
1091220 73.640625
1097575 73.640625
1098955 73.640625
1099310 73.640625
1101330 73.60625
1103610 73.60625
1104190 73.60625
1106355 73.60625
1109745 73.60625
1110225 73.60625
1117690 73.60625
1124230 73.60625
1126025 73.46875
1131240 73.46875
1131730 73.46875
1131790 73.46875
1137060 73.46875
1139830 73.46875
1139840 73.46875
1140860 73.46875
1142920 73.46875
1142935 73.46875
1145000 73.46875
1146455 73.46875
1147225 73.46875
1147870 73.46875
1149690 73.46875
1151220 72.9546875
1155120 72.9546875
1160305 72.9546875
1160875 72.9546875
1160965 72.9546875
1162180 72.9546875
1167355 72.9546875
1167430 72.9546875
1168165 72.9546875
1169050 72.9546875
1169690 72.9546875
1172295 72.9546875
1174590 72.9546875
1175300 72.275
1176080 72.275
1176550 72.275
1184535 72.275
1190240 72.275
1193080 72.275
1194075 72.275
1196695 72.275
1200475 72.1421875
1207660 72.1421875
1208150 72.1421875
1210325 72.1421875
1211710 72.1421875
1217860 72.1421875
1224505 72.1421875
1228575 72.1421875
1230255 72.1421875
1234700 72.1421875
1237540 72.1421875
1238475 72.1421875
1241465 72.1421875
1242180 72.1421875
1244805 72.1421875
1248385 72.1421875
1250125 72.1375
1250225 72.1375
1251215 72.1375
1254350 72.1375
1256600 72.1375
1258480 72.1375
1258985 72.1375
1259050 72.1375
1260335 72.1375
1261700 72.1375
1262930 72.1375
1264820 72.1375
1266005 72.1375
1267200 72.1375
1271475 72.1375
1272115 72.1375
1273010 72.1375
1273455 72.1375
1277315 71.75
1279980 71.75
1283655 71.75
1284730 71.75
1290315 71.75
1291040 71.75
1291755 71.75
1292845 71.75
1294800 71.75
1295560 71.75
1295615 71.75
1298685 71.75
1300180 71.43125
1301060 71.43125
1302660 71.43125
1304770 71.43125
1311630 71.43125
1312565 71.43125
1312640 71.43125
1312995 71.43125
1314270 71.43125
1314320 71.43125
1316505 71.43125
1319740 71.43125
1321065 71.43125
1324805 71.43125
1326365 71.43125
1328010 71.43125
1328475 71.43125
1330625 71.43125
1333295 71.43125
1338220 71.43125
1338250 71.43125
1339035 71.43125
1339045 71.43125
1339490 71.43125
1343440 71.43125
1343585 71.43125
1346365 71.43125
1351710 71.0171875
1352970 71.0171875
1354310 71.0171875
1355265 71.0171875
1355930 71.0171875
1357785 71.0171875
1360370 71.0171875
1361725 71.0171875
1363760 71.0171875
1364695 71.0171875
1368920 71.0171875
1377015 70.578125
1377030 70.578125
1377635 70.578125
1383840 70.578125
1387200 70.578125
1393740 70.578125
1394400 70.578125
1395245 70.578125
1397365 70.578125
1397410 70.578125
1397510 70.578125
1399455 70.578125
1402270 70.56875
1403095 70.56875
1404525 70.56875
1406390 70.56875
1407685 70.56875
1408515 70.56875
1411150 70.56875
1413255 70.56875
1418535 70.56875
1421520 70.56875
1424435 70.56875
1427690 70.565625
1431060 70.565625
1431175 70.565625
1433520 70.565625
1435680 70.565625
1435720 70.565625
1436935 70.565625
1443805 70.565625
1444690 70.565625
1446265 70.565625
1446825 70.565625
1449930 70.5625
1454085 70.5625
1465090 70.5625
1469135 70.5625
1470010 70.5625
1477045 70.55
1477870 70.55
1484410 70.55
1485930 70.55
1486500 70.55
1491760 70.55
1492070 70.55
1495845 70.55
1497430 70.55
1497745 70.55
1503575 70.55
1504170 70.55
1504970 70.55
1505145 70.55
1509150 70.55
1512160 70.55
1512475 70.55
1515335 70.55
1520120 70.55
1521400 70.55
1526525 70.55
1532535 70.55
1536355 70.55
1540625 70.55
1542060 70.55
1544280 70.55
1545050 70.55
1548740 70.55
1548825 70.55
1549585 70.55
1549725 70.55
1550535 70.55
1551475 70.55
1558200 70.55
1562940 70.55
1563780 70.55
1566690 70.55
1568685 70.55
1569535 70.55
1569740 70.55
1571460 70.55
1573905 70.55
1575850 70.55
1577950 70.55
1580985 70.55
1583980 70.55
1588490 70.55
1588535 70.55
1590035 70.55
1592050 70.55
1594145 70.55
1595560 70.55
1597180 70.55
1599710 70.5484375
1600070 70.5484375
1601415 70.5484375
1601960 70.5484375
1603720 70.5484375
1604295 70.5484375
1604640 70.5484375
1604780 70.5484375
1606920 70.5484375
1611945 70.5484375
1615045 70.5484375
1615250 70.5484375
1617640 70.5484375
1621525 70.5484375
1623450 70.5484375
1625315 70.5078125
1631900 70.5078125
1633385 70.5078125
1640195 70.5078125
1642720 70.5078125
1642725 70.5078125
1645255 70.5078125
1647715 70.5078125
1648555 70.5078125
1650240 70.1734375
1651660 70.1734375
1654850 70.1734375
1658195 70.1734375
1659750 70.1734375
1659885 70.1734375
1660640 70.1734375
1661295 70.1734375
1663605 70.1734375
1664035 70.1734375
1666290 70.1734375
1666630 70.1734375
1668820 70.1734375
1672215 70.1734375
1672245 70.1734375
1672390 70.1734375
1678135 69.9796875
1681920 69.9796875
1681965 69.9796875
1683625 69.9796875
1693545 69.9796875
1693685 69.9796875
1694455 69.9796875
1697730 69.9796875
1703580 69.3
1708810 69.3
1709040 69.3
1709105 69.3
1715330 69.3
1715750 69.3
1722520 69.3
1722670 69.3
1723610 69.3
1726400 69.0453125
1726965 69.0453125
1735560 69.0453125
1740925 69.0453125
1742710 69.0453125
1744625 69.0453125
1746775 69.0453125
1762805 69.0109375
1763145 69.0109375
1763170 69.0109375
1766030 69.0109375
1771260 69.0109375
1783475 68.6578125
1789625 68.6578125
1790420 68.6578125
1790815 68.6578125
1790985 68.6578125
1792915 68.6578125
1794040 68.6578125
1794925 68.6578125
1795420 68.6578125
1796915 68.15625
1797860 68.15625
1798955 68.15625
1801670 67.51875
1802250 67.51875
1803875 67.51875
1805610 67.51875
1809230 67.51875
1809645 67.51875
1811795 67.51875
1813745 67.51875
1815310 67.51875
1819255 67.51875
1821675 67.51875
1823855 67.51875
1823995 67.51875
1826840 67.4234375
1834345 67.4234375
1841585 67.4234375
1843610 67.4234375
1845470 67.4234375
1848530 67.4234375
1850010 67.3953125
1854190 67.3953125
1855630 67.3953125
1856265 67.3953125
1856580 67.3953125
1858640 67.3953125
1869140 67.3953125
1871940 67.3953125
1874280 67.3953125
1875260 67.3578125
1876290 67.3578125
1877750 67.3578125
1879145 67.3578125
1882440 67.3578125
1884830 67.3578125
1885560 67.3578125
1885680 67.3578125
1886415 67.3578125
1886860 67.3578125
1889630 67.3578125
1891160 67.3578125
1896435 67.3578125
1898675 67.33125
1899145 67.33125
1900935 67.3078125
1901910 67.3078125
1903240 67.3078125
1908265 67.3078125
1911045 67.3078125
1913180 67.3078125
1914560 67.3078125
1915605 67.3078125
1919130 67.3078125
1919185 67.3078125
1922130 67.3078125
1924055 67.3078125
1924620 67.3078125
1927540 67.296875
1928775 67.296875
1930600 67.296875
1931360 67.296875
1932285 67.296875
1937500 67.296875
1938235 67.296875
1938550 67.296875
1939915 67.296875
1940485 67.296875
1942460 67.296875
1944195 67.296875
1945875 67.296875
1948080 67.296875
1956110 67.296875
1961220 67.296875
1965965 67.296875
1967040 67.296875
1971795 67.296875
1974570 67.296875
1975870 67.29375
1978035 67.29375
1980205 67.29375
1980795 67.29375
1984300 67.29375
1987555 67.29375
1988855 67.29375
1989680 67.29375
1990985 67.29375
1991035 67.29375
1996415 67.29375
2000995 67.290625
2002515 67.290625
2003175 67.290625
2008195 67.290625
2008430 67.290625
2015555 67.290625
2016460 67.290625
2017045 67.290625
2017430 67.290625
2021010 67.290625
2023160 67.290625
2023895 67.290625
2025065 67.290625
2028800 67.290625
2029240 67.290625
2032475 67.290625
2036330 67.290625
2037590 67.290625
2039750 67.290625
2042755 67.290625
2049850 67.290625
2055910 67.290625
2056290 67.290625
2060385 67.290625
2063415 67.290625
2065730 67.290625
2068055 67.290625
2069820 67.290625
2075085 67.2859375
2075685 67.2859375
2078185 67.2859375
2080530 67.2859375
2085815 67.2859375
2087825 67.2859375
2089005 67.2859375
2089045 67.2859375
2091560 67.2859375
2095940 67.2859375
2099345 67.2859375
2102625 67.2859375
2107490 67.2859375
2107660 67.2859375
2111245 67.2859375
2112195 67.2859375
2112470 67.2859375
2112535 67.2859375
2116105 67.2859375
2117340 67.2859375
2118765 67.2859375
2125160 67.2578125
2125185 67.2578125
2128095 67.2578125
2132495 67.2578125
2132880 67.2578125
2135285 67.2578125
2137435 67.2578125
2140865 67.2578125
2141300 67.2578125
2142875 67.2578125
2150385 67.1859375
2150660 67.1859375
2150980 67.1859375
2152445 67.1859375
2153660 67.1859375
2159375 67.1859375
2160365 67.1859375
2160965 67.1859375
2161355 67.1859375
2168060 67.1859375
2168880 67.1859375
2173350 67.1859375
2174555 67.1859375
2174600 67.1859375
2177615 67.1859375
2178150 67.1859375
2178360 67.1859375
2181055 67.1859375
2185090 67.1859375
2186985 67.1859375
2191545 67.1859375
2192140 67.1859375
2194130 67.1859375
2198885 67.1859375
2199000 67.1859375
2205400 67.140625
2209450 67.140625
2211255 67.140625
2211450 67.140625
2212685 67.140625
2213185 67.140625
2213275 67.140625
2215275 67.140625
2218930 67.140625
2219200 67.140625
2224380 67.140625
2226760 67.13125
2226800 67.13125
2227100 67.13125
2229995 67.13125
2236900 67.13125
2237520 67.13125
2241975 67.13125
2242280 67.13125
2243205 67.13125
2244990 67.13125
2246110 67.13125
2246545 67.13125
2247160 67.13125
2252165 67.13125
2252330 67.13125
2259635 67.13125
2261265 67.13125
2262745 67.13125
2265370 67.13125
2266105 67.13125
2266175 67.13125
2267920 67.13125
2270100 67.13125
2271995 67.13125
2278380 67.13125
2287085 67.13125
2289685 67.13125
2289795 67.13125
2292250 67.13125
2292515 67.13125
2297515 67.13125
2299910 67.13125
2301835 67.1296875
2305025 67.1296875
2306610 67.1296875
2308320 67.1296875
2312355 67.1296875
2316580 67.1296875
2316670 67.1296875
2317715 67.1296875
2322620 67.1296875
2330600 67.1296875
};
\addlegendentry{Data}
\end{axis}

\end{tikzpicture}

%% file: figures/torcs3.tikz
\begin{tikzpicture}[scale=0.8]

\begin{axis}[
legend cell align={left},
legend style={fill opacity=0.8, draw opacity=1, text opacity=1, at={(0.03,0.03)}, anchor=south west, draw=white!80!black},
log basis x={10},
tick align=outside,
tick pos=left,
x grid style={white!69.0196078431373!black},
xlabel={Num. of samples},
xmin=523.763060204103, xmax=2500000,
xmode=log,
xtick style={color=black},
xtick={10,100,1000,10000,100000,1000000,10000000,100000000},
xticklabels={\(\displaystyle {10^{1}}\),\(\displaystyle {10^{2}}\),\(\displaystyle {10^{3}}\),\(\displaystyle {10^{4}}\),\(\displaystyle {10^{5}}\),\(\displaystyle {10^{6}}\),\(\displaystyle {10^{7}}\),\(\displaystyle {10^{8}}\)},
y grid style={white!69.0196078431373!black},
ylabel={Best lap time (s)},
ymin=70, ymax=170,
ytick style={color=black}
]
\addplot [semithick, blue]
table {%
784 150
1509 138.6
2142 120.6
2729 111.6
3279 104.4
3797 98
5748 98
6269 98
7035 98
7550 97.4
7805 97.4
8298 93
8762 87.2
9249 87.2
9702 85
10006 85
11459 85
12851 85
14724 85
15254 85
15754 85
16231 85
16919 85
17397 85
17853 85
18129 85
18812 85
20898 85
23378 85
23948 85
24485 85
24775 85
25060 85
25544 85
26071 85
28026 85
28557 85
28843 85
29353 85
29843 85
30139 85
30841 85
32795 85
33317 85
33817 85
36229 85
36739 85
37016 85
37692 85
39096 85
39657 85
40188 85
41546 85
42064 85
42559 85
42825 85
43302 85
43589 85
45549 85
46060 85
46549 85
46841 85
47323 85
47787 85
48075 85
48837 85
50184 85
50685 85
51944 85
53639 85
55334 85
55842 85
56523 85
57784 85
58297 85
58567 85
59068 85
60315 85
60818 85
61110 85
61415 85
61910 85
63887 85
66391 85
66953 85
67487 85
69891 85
70425 85
71110 85
73377 85
73929 85
74189 85
74715 85
75003 85
75515 85
77205 85
79684 85
80228 85
82560 85
83110 85
};
\addlegendentry{Ours}
\addplot [semithick, blue, forget plot]
table {%
83110 85
2500000 85
};
\addplot [semithick, green!50!black]
table {%
1 150
2500000 150
};
\addlegendentry{Mechanism}
\addplot [semithick, red]
table {%
840 300
2375 300
2685 300
2935 300
3240 300
7185 300
8690 300
11540 300
11950 300
12435 300
16235 300
22465 300
25675 300
29540 300
33500 300
33560 300
34835 300
37320 300
38995 300
39170 300
42495 300
46305 300
47345 300
52000 300
54530 300
54835 300
57665 300
59850 300
64390 300
65175 300
71475 300
73345 300
73900 300
80585 300
84290 300
86590 300
87720 300
97660 300
98205 300
100315 300
102700 300
102895 300
104280 300
105405 300
105725 300
106600 300
111655 300
111795 300
116435 300
116810 300
117140 300
118355 300
118510 300
120740 300
121615 300
123640 300
124675 300
126270 300
133990 300
144855 300
144985 300
149150 300
152355 300
153065 300
153535 300
157035 300
157220 300
159685 300
162180 300
165910 300
168690 300
168750 300
168930 300
169200 300
169385 300
169570 300
171815 300
172630 300
173000 300
177420 300
179770 300
180190 300
183150 300
183300 300
184255 300
186015 300
187450 300
188825 300
189385 300
193605 300
194205 300
196485 300
198515 300
203970 300
206360 300
210675 300
212345 300
218200 300
218560 300
219030 300
219970 300
221005 300
221260 300
221525 300
224085 300
224825 300
226865 300
227540 300
228295 300
231675 300
232070 300
238725 300
239855 300
240910 300
246365 300
250285 300
254100 300
255675 300
256140 300
258995 300
264535 300
265465 300
267345 300
269315 300
274040 300
281690 300
282610 300
284105 300
293600 300
293735 300
293820 300
300900 295.1921875
300930 295.1921875
302380 295.1921875
310700 295.1921875
312740 295.1921875
318360 295.1921875
322610 295.1921875
328220 279.15
333940 279.15
335110 279.15
337205 279.15
345930 279.15
347280 279.15
348600 279.15
349580 279.15
352315 274.2515625
352905 274.2515625
353285 274.2515625
354985 274.2515625
357145 274.2515625
359535 274.2515625
359790 274.2515625
359900 274.2515625
360065 274.2515625
363620 274.2515625
365050 274.2515625
377440 257.9921875
381095 257.9921875
384535 257.9921875
384570 257.9921875
391045 257.9921875
394605 257.9921875
394870 257.9921875
395030 257.9921875
401955 119.2453125
403070 119.2453125
404985 119.2453125
405080 119.2453125
416340 119.2453125
422050 119.2453125
423010 119.2453125
425450 119.2453125
430075 119.2453125
433905 119.2453125
434330 119.2453125
435045 119.2453125
436970 119.2453125
440690 119.2453125
443375 119.2453125
446740 119.2453125
446820 119.2453125
449870 119.2453125
451190 119.2453125
456505 119.2453125
458125 119.2453125
458240 119.2453125
460605 119.2453125
466120 119.2453125
467360 119.2453125
469510 119.2453125
473925 119.2453125
475290 119.2453125
476890 119.2453125
477405 119.2453125
481510 119.2453125
489935 119.2453125
492375 119.2453125
495280 119.2453125
499425 119.2453125
504295 116.0703125
505365 116.0703125
506680 116.0703125
509620 116.0703125
509720 116.0703125
514405 116.0703125
514560 116.0703125
514835 116.0703125
515010 116.0703125
515060 116.0703125
519740 116.0703125
521385 116.0703125
521675 116.0703125
524485 116.0703125
524625 116.0703125
525470 113.59375
527175 113.59375
528875 113.59375
531340 113.59375
532835 113.59375
539190 113.59375
542935 113.59375
551125 111.4515625
552095 111.4515625
553210 111.4515625
556620 111.4515625
558495 111.4515625
564515 111.4515625
566615 111.4515625
568710 111.4515625
569905 111.4515625
571465 111.4515625
575485 109.69375
577535 109.69375
577595 109.69375
579395 109.69375
583370 109.69375
583560 109.69375
583920 109.69375
585240 109.69375
588180 109.69375
589130 109.69375
589710 109.69375
590955 109.69375
605965 104.640625
605970 104.640625
606665 104.640625
607020 104.640625
609180 104.640625
611030 104.640625
613115 104.640625
613175 104.640625
615345 104.640625
618335 104.640625
620990 104.640625
626455 103.25
637715 103.25
638190 103.25
645735 103.25
646595 103.25
647035 103.25
647515 103.25
650360 98.3140625
658405 98.3140625
659385 98.3140625
669305 98.3140625
671885 98.3140625
671950 98.3140625
672375 98.3140625
674390 98.3140625
678080 96.09375
681050 96.09375
683370 96.09375
683395 96.09375
685280 96.09375
687830 96.09375
694700 96.09375
696295 96.09375
697020 96.09375
697250 96.09375
699880 96.09375
700080 92.925
700230 92.925
700280 92.925
700660 92.925
703755 92.925
704950 92.925
706115 92.925
709450 92.925
710985 92.925
710990 92.925
713760 92.925
714465 92.925
715935 92.925
719595 92.925
721135 92.925
721430 92.925
722025 92.925
731565 91.934375
733485 91.934375
737460 91.934375
740950 91.934375
743025 91.934375
743370 91.934375
743390 91.934375
747215 91.934375
748420 91.934375
749435 91.934375
753795 91.709375
754320 91.709375
755215 91.709375
755285 91.709375
755465 91.709375
755670 91.709375
756140 91.709375
756385 91.709375
757720 91.709375
762620 91.709375
763835 91.709375
775950 91.1078125
777925 91.1078125
778605 91.1078125
778680 91.1078125
783265 91.1078125
785325 91.1078125
785655 91.1078125
787155 91.1078125
788120 91.1078125
797045 91.1078125
798900 91.1078125
801015 90.675
801265 90.675
802125 90.675
803545 90.675
805800 90.675
805940 90.675
806825 90.675
809755 90.675
811035 90.675
812920 90.675
813950 90.675
816485 90.675
823115 90.675
827320 90.675
827990 90.675
829005 90.675
831495 90.675
836175 90.675
836985 90.675
838225 90.675
840190 90.675
846620 90.675
847775 90.675
849370 90.675
849820 90.675
850945 90.675
852480 90.675
856065 90.675
858105 90.675
858300 90.675
858350 90.675
864865 90.675
865470 90.675
865875 90.675
867785 90.675
868730 90.675
872075 90.675
873240 90.675
882920 90.603125
888890 90.603125
890965 90.603125
893310 90.603125
893510 90.603125
895985 90.603125
899835 90.603125
901580 90.2421875
905250 90.2421875
905545 90.2421875
907320 90.2421875
908185 90.2421875
910440 90.2421875
914180 90.2421875
914725 90.2421875
914750 90.2421875
916125 90.2421875
916360 90.2421875
917435 90.2421875
917995 90.2421875
923730 90.2421875
925295 89.8640625
930030 89.8640625
931165 89.8640625
933445 89.8640625
935455 89.8640625
935705 89.8640625
938375 89.8640625
939720 89.8640625
939950 89.8640625
946300 89.8640625
946615 89.8640625
949670 89.8640625
951695 89.3484375
952215 89.3484375
961080 89.3484375
973110 89.3484375
976490 89.1796875
977005 89.1796875
978100 89.1796875
983635 89.1796875
983885 89.1796875
991010 89.1796875
994260 89.1796875
994280 89.1796875
996440 89.1796875
998315 89.1796875
1001060 88.3296875
1007010 88.3296875
1008930 88.3296875
1009605 88.3296875
1014850 88.3296875
1023770 88.3296875
1027165 86.540625
1030555 86.540625
1035255 86.540625
1038050 86.540625
1043745 86.540625
1044675 86.540625
1049945 86.540625
1050960 85.746875
1056210 85.746875
1058485 85.746875
1065500 85.746875
1065885 85.746875
1069340 85.746875
1070430 85.746875
1070605 85.746875
1073055 85.746875
1073640 85.746875
1076655 85.175
1079770 85.175
1080120 85.175
1080265 85.175
1089340 85.175
1091105 85.175
1091220 85.175
1097575 85.175
1098955 85.175
1099310 85.175
1101330 85.0578125
1104190 85.0578125
1106355 85.0578125
1109745 85.0578125
1110225 85.0578125
1117690 85.0578125
1124230 85.0578125
1126025 84.959375
1131240 84.959375
1131730 84.959375
1131790 84.959375
1137060 84.959375
1139830 84.959375
1139840 84.959375
1140860 84.959375
1142920 84.959375
1142935 84.959375
1145000 84.959375
1146455 84.959375
1147225 84.959375
1147870 84.959375
1151220 84.8359375
1155120 84.8359375
1160305 84.8359375
1160875 84.8359375
1160965 84.8359375
1162180 84.8359375
1167355 84.8359375
1167430 84.8359375
1168165 84.8359375
1169050 84.8359375
1169690 84.8359375
1172295 84.8359375
1174590 84.8359375
1175300 84.6109375
1176080 84.6109375
1176550 84.6109375
1184535 84.6109375
1190240 84.6109375
1193080 84.6109375
1194075 84.6109375
1196695 84.6109375
1200475 84.5890625
1207660 84.5890625
1208150 84.5890625
1210325 84.5890625
1211710 84.5890625
1217860 84.5890625
1224505 84.5890625
1228575 84.5890625
1230255 84.5890625
1234700 84.5890625
1237540 84.5890625
1238475 84.5890625
1241465 84.5890625
1242180 84.5890625
1244805 84.5890625
1248385 84.5890625
1250125 84.5890625
1251215 84.5890625
1254350 84.5890625
1256600 84.5890625
1258480 84.5890625
1258985 84.5890625
1259050 84.5890625
1260335 84.5890625
1261700 84.5890625
1262930 84.5890625
1266005 84.5890625
1267200 84.5890625
1271475 84.5890625
1272115 84.5890625
1273010 84.5890625
1273455 84.5890625
1277315 84.5890625
1279980 84.5890625
1283655 84.5890625
1284730 84.5890625
1290315 84.5890625
1291040 84.5890625
1291755 84.5890625
1292845 84.5890625
1294800 84.5890625
1295560 84.5890625
1295615 84.5890625
1298685 84.5890625
1300180 84.575
1301060 84.575
1302660 84.575
1304770 84.575
1311630 84.575
1312565 84.575
1312640 84.575
1312995 84.575
1314270 84.575
1316505 84.575
1319740 84.575
1321065 84.575
1324805 84.575
1326365 84.571875
1328010 84.571875
1328475 84.571875
1330625 84.571875
1333295 84.571875
1338220 84.571875
1339035 84.571875
1339045 84.571875
1339490 84.571875
1343440 84.571875
1343585 84.571875
1346365 84.571875
1351710 84.571875
1352970 84.571875
1354310 84.571875
1355265 84.571875
1355930 84.571875
1357785 84.571875
1360370 84.571875
1361725 84.571875
1363760 84.571875
1364695 84.571875
1368920 84.571875
1377015 84.5453125
1377030 84.5453125
1377635 84.5453125
1383840 84.5453125
1387200 84.5453125
1393740 84.5453125
1394400 84.5453125
1395245 84.5453125
1397365 84.5453125
1397410 84.5453125
1397510 84.5453125
1399455 84.5453125
1402270 84.375
1403095 84.375
1404525 84.375
1406390 84.375
1407685 84.375
1408515 84.375
1411150 84.375
1413255 84.375
1418535 84.375
1421520 84.375
1424435 84.375
1427690 84.31875
1431175 84.31875
1433520 84.31875
1435680 84.31875
1435720 84.31875
1436935 84.31875
1443805 84.31875
1444690 84.31875
1446265 84.31875
1446825 84.31875
1449930 84.31875
1454085 84.08125
1465090 84.08125
1469135 84.08125
1470010 84.08125
1477045 83.6703125
1477870 83.6703125
1484410 83.6703125
1491760 83.6703125
1492070 83.6703125
1495845 83.6703125
1497430 83.6703125
1503575 83.4875
1504170 83.4875
1504970 83.4875
1505145 83.4875
1509150 83.4875
1512160 83.4875
1512475 83.4875
1515335 83.4875
1520120 83.4875
1521400 83.4875
1526525 83.3921875
1532535 83.3921875
1536355 83.3921875
1542060 83.3921875
1544280 83.3921875
1545050 83.3921875
1548740 83.3921875
1548825 83.3921875
1549585 83.3921875
1549725 83.3921875
1550535 83.2875
1551475 83.2875
1558200 83.2875
1562940 83.2875
1563780 83.2875
1568685 83.2875
1569535 83.2875
1569740 83.2875
1571460 83.2875
1573905 83.2875
1575850 83.053125
1577950 83.053125
1580985 83.053125
1583980 83.053125
1588490 83.053125
1590035 83.053125
1592050 83.053125
1594145 83.053125
1595560 83.053125
1597180 83.053125
1599710 83.053125
1600070 82.61875
1601415 82.61875
1601960 82.61875
1603720 82.61875
1604295 82.61875
1604640 82.61875
1604780 82.61875
1606920 82.61875
1611945 82.61875
1615045 82.61875
1615250 82.61875
1617640 82.61875
1621525 82.61875
1623450 82.61875
1625315 81.6703125
1631900 81.6703125
1633385 81.6703125
1640195 81.6703125
1642720 81.6703125
1642725 81.6703125
1645255 81.6703125
1648555 81.6703125
1650240 81.359375
1651660 81.359375
1654850 81.359375
1658195 81.359375
1659750 81.359375
1659885 81.359375
1660640 81.359375
1661295 81.359375
1663605 81.359375
1664035 81.359375
1666290 81.359375
1666630 81.359375
1668820 81.359375
1672215 81.359375
1672245 81.359375
1672390 81.359375
1678135 81.059375
1681920 81.059375
1681965 81.059375
1683625 81.059375
1693545 81.059375
1693685 81.059375
1694455 81.059375
1697730 81.059375
1708810 80.2625
1709040 80.2625
1709105 80.2625
1715330 80.2625
1715750 80.2625
1722520 80.2625
1722670 80.2625
1723610 80.2625
1726400 79.653125
1726965 79.653125
1735560 79.653125
1740925 79.653125
1742710 79.653125
1744625 79.653125
1746775 79.653125
1762805 79.2453125
1763145 79.2453125
1763170 79.2453125
1766030 79.2453125
1771260 79.2453125
1783475 79.1546875
1789625 79.1546875
1790420 79.1546875
1790815 79.1546875
1790985 79.1546875
1792915 79.1546875
1794040 79.1546875
1796915 79.1546875
1798955 79.1546875
1801670 79.025
1802250 79.025
1803875 79.025
1805610 79.025
1809230 79.025
1809645 79.025
1811795 79.025
1813745 79.025
1815310 79.025
1821675 79.025
1823855 79.025
1823995 79.025
1826840 78.9640625
1834345 78.9640625
1841585 78.9640625
1843610 78.9640625
1845470 78.9640625
1848530 78.9640625
1850010 78.8875
1854190 78.8875
1855630 78.8875
1856265 78.8875
1856580 78.8875
1858640 78.8875
1869140 78.8875
1871940 78.8875
1874280 78.8875
1875260 78.753125
1876290 78.753125
1877750 78.753125
1879145 78.753125
1882440 78.753125
1884830 78.753125
1885560 78.753125
1885680 78.753125
1886415 78.753125
1886860 78.753125
1889630 78.753125
1891160 78.753125
1896435 78.753125
1898675 78.753125
1899145 78.753125
1900935 78.63125
1901910 78.63125
1903240 78.63125
1908265 78.63125
1911045 78.63125
1913180 78.63125
1914560 78.63125
1915605 78.63125
1919130 78.63125
1919185 78.63125
1924055 78.63125
1924620 78.63125
1927540 78.4859375
1928775 78.4859375
1930600 78.4859375
1931360 78.4859375
1932285 78.4859375
1937500 78.4859375
1938235 78.4859375
1938550 78.4859375
1939915 78.4859375
1940485 78.4859375
1942460 78.4859375
1944195 78.4859375
1945875 78.4859375
1948080 78.4859375
1956110 78.128125
1961220 78.128125
1965965 78.128125
1967040 78.128125
1971795 78.128125
1974570 78.128125
1975870 77.75625
1978035 77.75625
1980205 77.75625
1980795 77.75625
1984300 77.75625
1987555 77.75625
1988855 77.75625
1989680 77.75625
1990985 77.75625
1991035 77.75625
1996415 77.75625
2000995 77.5890625
2002515 77.5890625
2003175 77.5890625
2015555 77.5890625
2016460 77.5890625
2017045 77.5890625
2017430 77.5890625
2021010 77.5890625
2023895 77.5890625
2028800 77.4140625
2029240 77.4140625
2037590 77.4140625
2039750 77.4140625
2042755 77.4140625
2049850 77.4140625
2055910 77.3609375
2056290 77.3609375
2060385 77.3609375
2063415 77.3609375
2065730 77.3609375
2068055 77.3609375
2069820 77.3609375
2075085 77.35
2075685 77.35
2078185 77.35
2080530 77.35
2085815 77.35
2087825 77.35
2089005 77.35
2089045 77.35
2091560 77.35
2095940 77.35
2099345 77.35
2102625 77.315625
2107490 77.315625
2107660 77.315625
2111245 77.315625
2112195 77.315625
2112470 77.315625
2112535 77.315625
2117340 77.315625
2118765 77.315625
2125160 77.3125
2125185 77.3125
2128095 77.3125
2132495 77.3125
2132880 77.3125
2135285 77.3125
2137435 77.3125
2140865 77.3125
2141300 77.3125
2142875 77.3125
2150385 77.3125
2150660 77.3125
2150980 77.3125
2152445 77.3125
2153660 77.3125
2159375 77.3125
2160365 77.3125
2160965 77.3125
2161355 77.3125
2168060 77.3125
2168880 77.3125
2173350 77.3125
2174555 77.3125
2174600 77.3125
2177615 77.3125
2178150 77.3125
2178360 77.3125
2181055 77.3125
2186985 77.3125
2191545 77.3125
2192140 77.3125
2194130 77.3125
2199000 77.3125
2205400 77.3015625
2209450 77.3015625
2211255 77.3015625
2211450 77.3015625
2212685 77.3015625
2213185 77.3015625
2213275 77.3015625
2215275 77.3015625
2218930 77.3015625
2219200 77.3015625
2224380 77.3015625
2226760 77.3015625
2227100 77.3015625
2229995 77.3015625
2237520 77.3015625
2241975 77.3015625
2242280 77.3015625
2243205 77.3015625
2244990 77.3015625
2246110 77.3015625
2247160 77.3015625
2252165 77.3015625
2252330 77.3015625
2259635 77.3015625
2261265 77.3015625
2262745 77.3015625
2265370 77.3015625
2266105 77.3015625
2266175 77.3015625
2267920 77.3015625
2270100 77.3015625
2271995 77.3015625
2278380 77.296875
2287085 77.296875
2289795 77.296875
2292250 77.296875
2292515 77.296875
2297515 77.296875
2299910 77.296875
2301835 77.296875
2305025 77.296875
2306610 77.296875
2308320 77.296875
2312355 77.296875
2316580 77.296875
2316670 77.296875
2322620 77.296875
2331355 77.296875
2334435 77.296875
2334805 77.296875
2335310 77.296875
2339405 77.296875
2342900 77.296875
2343005 77.296875
2345105 77.296875
2346970 77.296875
2349670 77.296875
2352845 77.2828125
2359675 77.2828125
2360805 77.2828125
2362465 77.2828125
2363245 77.2828125
2363775 77.2828125
2364330 77.2828125
2377410 77.203125
2381630 77.203125
2382200 77.203125
2383320 77.203125
2384490 77.203125
2385615 77.203125
2385665 77.203125
2387685 77.203125
2392445 77.203125
2393690 77.203125
2393815 77.203125
2395200 77.203125
2395815 77.203125
2398980 77.203125
2402035 77.121875
2404175 77.121875
2405730 77.121875
2406470 77.121875
2406540 77.121875
2406930 77.121875
2423365 77.121875
2424705 77.121875
2424850 77.121875
2425400 77.0890625
2432635 77.0890625
2434510 77.0890625
2442170 77.0890625
2444195 77.0890625
2451360 77.084375
2452320 77.084375
2452580 77.084375
2455640 77.084375
2461365 77.084375
2462310 77.084375
2471025 77.084375
2471425 77.084375
2473775 77.084375
2479540 77.071875
2482210 77.071875
2488255 77.071875
2493330 77.071875
2494415 77.071875
2495100 77.071875
2495595 77.071875
2500000 77.071875
};
\addlegendentry{Data}
\end{axis}

\end{tikzpicture}